\documentclass[10pt,twocolumn,letterpaper]{article}

\usepackage{cvpr}              
\usepackage{makecell}
\definecolor{cvprblue}{rgb}{0.21,0.49,0.74}
\usepackage[pagebackref,breaklinks,colorlinks,allcolors=cvprblue]{hyperref}
\usepackage[acronym]{glossaries}
\usepackage{multirow}
\def\paperID{*****} 
\def\confName{CVPR}
\def\confYear{2025}
\makenoidxglossaries%
\newacronym{fta}{FTA}{Fehlerbaumanalyse}%
\newacronym{ctan}{CTAN}{Comprehensive-TeX-Archive-Network}%
\newacronym{oem}{OEM}{Original Equipment Manufacturer}%
\newacronym{pkw}{PKW}{Personenkraftwagen}%
\newacronym{suv}{SUV}{Sport Utility Vehicle}%
\newacronym[description={Antiblockiersystem (engl. Anti-lock Braking System)}, \glslongpluralkey={Antiblockiersysteme},\glsshortpluralkey={ABS}]{abs}{ABS}{Antiblockiersystem}%
\newacronym{asr}{ASR}{Antischlupfregelung}%
\newacronym[plural={LLMs}]{llm}{LLM}{Large Language Model}
\newacronym{ai}{AI}{artificial intelligence}
\newacronym{ml}{ML}{machine learning}
\newacronym{lvlm}{LVLM}{Large Vision Language Model}
\newacronym{mllm}{MLLM}{Multimodal Large Language Model}
\newacronym{vqa}{VQA}{visual question answering}
\newacronym{vlm}{VLM}{Visual language model}
\newacronym{bev}{BEV}{birds-eye view}
\newacronym{mlp}{MLP}{Multi-Layer Perceptron}
\newacronym{ffn}{FFN}{Feed-Forward Network}
\newacronym{sam}{SAM}{Segment Anything}
\newacronym{yolo}{YOLO}{You Only Look Once}
\newacronym{cnn}{CNN}{Convolutional Neural Network}
\newacronym{vit}{ViT}{Vision Transformer}
\newacronym{clip}{CLIP}{Contrastive Language-Image Pre-Training}
\newacronym{llavav}{V3LMA-L}{LLaVA-Video-7B-Qwen2+Qwen2.5-7B-Instruct}
\newacronym{qwen14}{V3LMA-Q}{Qwen2-VL-7B-Instruct+Qwen2.5-7B-Instruct}
\newacronym{qwen35}{V3LMA-Q-mini}{Qwen2-VL-2B-Instruct+Qwen2.5-1.5B-Instruct}
\newacronym{velma}{V3LMA}{Visual 3D-enhanced Language Model for Autonomous Driving}
\title{V3LMA: Visual 3D-enhanced Language Model for Autonomous Driving}

\author{Jannik Lübberstedt\thanks{Equal contribution}\text{ }$^{1}$ \quad Esteban Rivera\footnotemark[1]\text{ }$^{1}$$^{2}$\thanks{Corresponding author. \newline
The research was partially funded by the Bavarian Research Foundation
within the project ”Data-Enabled Autonomous Driving”, and through basic
research funds from the Institute for Automotive Technology} \quad Nico Uhlemann$^{1}$$^{2}$ \quad Markus Lienkamp$^{1}$$^{2}$\\
 $^1$Technical University of Munich (TUM)\\
$^2$Munich Institute of Robotics and Machine Intelligence (MIRMI) \\
{\tt\small firstname.lastname@tum.de}
}

\begin{document}
\maketitle
\begin{abstract}
\Glspl{lvlm} have shown strong capabilities in understanding and analyzing visual scenes across various domains. However, in the context of autonomous driving, their limited comprehension of 3D environments restricts their effectiveness in achieving a complete and safe understanding of dynamic surroundings. To address this, we introduce V3LMA, a novel approach that enhances 3D scene understanding by integrating \glspl{llm} with \glspl{lvlm}. V3LMA leverages textual descriptions generated from object detections and video inputs, significantly boosting performance without requiring fine-tuning. Through a dedicated preprocessing pipeline that extracts 3D object data, our method improves situational awareness and decision-making in complex traffic scenarios, achieving a score of 0.56 on the LingoQA benchmark. We further explore different fusion strategies and token combinations with the goal of advancing the interpretation of traffic scenes, ultimately enabling safer autonomous driving systems.
\end{abstract}    
\section{Introduction}
\label{sec:introduction}


For autonomous agents to navigate effectively in intricate scenarios, complete environmental understanding and interpretation are crucial. As a result, autonomous driving continues to be an unresolved research challenge, with much of the focus directed toward the fields of computer vision and natural language processing. An accurate understanding of complex scenes is essential for the development of autonomous vehicles, traffic monitoring systems, and smart city infrastructure. It involves not only detecting and interpreting objects, but also understanding the intricate relationships between them, the context, and the continuous changes in traffic dynamics.

\begin{figure}
    \centering
    \includegraphics[width=1\linewidth]{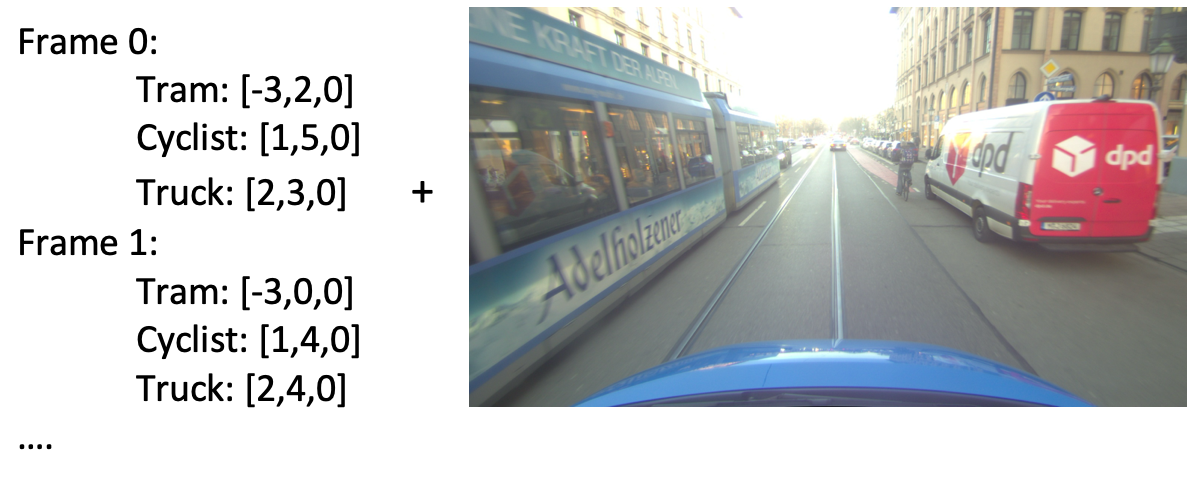}
    \caption{3D information is essential for accurate and safe scene understanding in autonomous driving. While \glspl{lvlm} can extract rich semantic information from images, they struggle to capture 3D spatial cues. To address this limitation, we propose enhancing visual scene understanding by incorporating textual descriptions of 3D detections. These descriptions are processed by an \gls{llm}, which is better equipped to handle textual input than an \gls{lvlm}.}
    \label{fig:example}
\end{figure}

\glspl{llm} have recently shown impressive capabilities in understanding and reasoning over textual representations of the natural world. These strengths extend to multimodal settings, where \glspl{llm} are augmented with visual inputs. Such models excel at extracting information from images when guided by textual queries and can even reason about visual scenes and predict the outcomes of actions depicted in them \cite{openai2024gpt4technicalreport} \cite{liu2023visualinstructiontuning}. However, these models are constrained by the data they are trained on, meaning their performance can significantly decline in specialized fields or when faced with unfamiliar data formats. To improve their reasoning abilities in a specific domain, these models are often fine-tuned using domain-specific data. Notable examples include medicine and biology, with ongoing efforts to fine-tune models to reason effectively in driving scenarios \cite{moor2023medflamingomultimodalmedicalfewshot}\cite{Hartsock_2024}\cite{zhang2025biomedclipmultimodalbiomedicalfoundation}\cite{tian2024drivevlmconvergenceautonomousdriving}. Yet, such fine-tuning demands considerable computational resources and access to high-quality, labeled datasets.

Recent developments in \glspl{mllm}, which integrate inputs from various sensory modalities such as images, video, text, and sensor data, have opened new opportunities for improving scene understanding in traffic environments \cite{guo2023pointbindpointllmaligning}\cite{Xu.31.08.2023}. These models are capable of fusing diverse input types to generate context-aware insights that can adapt to real-time, dynamic scenarios. Nevertheless, challenges remain as  \glspl{mllm} often struggle to understand the relationships between heterogeneous data types and interpret ambiguous or incomplete inputs, especially in complex traffic scenes.

In this work, we explore the use of video-based \glspl{lvlm} in combination with \glspl{llm} for scene understanding in autonomous driving. Our objective is to advance the capabilities of \gls{ai} systems in interpreting and responding to the diverse and unpredictable nature of traffic. By fusing visual and textual data, we aim to enhance situational awareness, support more informed decision-making, and contribute to the development of safer, more efficient transportation systems. Previous work has explored the use of textual representations alongside visual input to reduce hallucinations in \glspl{lvlm} outputs, often relying on fine-tuning to ensure accurate reasoning about the textual input \cite{tian2024drivevlmconvergenceautonomousdriving}. 

In contrast, we introduce \gls{velma}, a novel approach that integrates textual descriptions of object detections with visual video inputs into a combined \glspl{llm}-\gls{lvlm} architecture. Notably, while significantly boosting the performance of the base models, no fine-tuning is required. The textual detections are generated by a pre-processing pipeline that includes object detection and tracking modules. This pipeline is modular and easily adaptable, allowing for seamless integration of alternative models and modalities. Furthermore, by incorporating 3D detection information, \gls{velma} enables a deeper understanding of spatial relationships among traffic participants, contributing to a safer and more precise interpretation of complex traffic scenes. To summarize, the main contributions of this work are:
\begin{itemize}
    \item Proposal and evaluation of \gls{velma}, a novel method that combines the strengths of \glspl{llm} and \glspl{lvlm} to enhance 3D scene understanding in traffic scenarios—without requiring model training or fine-tuning.
    \item Design of a modular preprocessing pipeline that converts 3D object detections into textual input suitable for \glspl{llm}, enabling flexible integration of different detection and tracking models.
    \item A systematic study of fusion strategies for combining \glspl{lvlm} and \glspl{llm}, analyzing the impact of fusion layers, weight distributions, and token configurations.
\end{itemize}

\section{Related work}
\label{sec:relatedwork}
\subsection{Datasets}
To train and evaluate scene understanding models for autonomous driving, several datasets are available. Some of them extend existent datasets with novel captions describing the scenes \cite{kim2018textualexplanationsselfdrivingvehicles, inoue2023nuscenesmqaintegratedevaluationcaptions, wu2023languagepromptautonomousdriving, 10588373, inoue2023nuscenesmqaintegratedevaluationcaptions, Deruyttere_2019}. Besides, there are datasets which present question and answer pairs closely related to the driving task; some focus on image-level granularity  \cite{nie2024reason2driveinterpretablechainbasedreasoning, tencent2023maplm, qian2024nuscenesqamultimodalvisualquestion} or in video-level captioning \cite{xu2021sutdtrafficqaquestionansweringbenchmark, SungYeonPark.,marcu2024lingoqavisualquestionanswering}. A detailed description is given in appendix \Cref{table:language_datasets}. The only available dataset with Video annotations was LingoQA\cite{marcu2024lingoqavisualquestionanswering}, which we then use for our evaluation. It comprises approximately 28,000 unique short video scenarios captured in central London, accompanied by 419,000 question-answer pairs, facilitating the assessment of vision-language models in understanding and reasoning about real-world driving situations. Additionally, they introduce Lingo-Judge, a learned text classifier that serves as an evaluation metric, achieving a Spearman correlation coefficient of 0.95 with human evaluations, thereby providing a reliable measure of model performance.
\subsection{Scene understanding models}
\label{sec:understanding}
The authors of LingoQA published a baseline model alongside its dataset. This model uses a similar way of incorporating visual information into the \gls{llm} as LLaVA. Instead of using a linear projection layer to project the \gls{clip} features into the language feature space, it uses Querying Transformer (Q-Former) modules to retrieve a constant number of tokens per image. Similarly, the WiseAD model \cite{zhang2024wiseadknowledgeaugmentedendtoend} utilizes a mobile vision-language model architecture using a frozen \gls{clip} model, a learnable projector layer and the MobileLLaMA \cite{chu2024mobilevlmv2fasterstronger} language model. Despite its small size of around 2B parameters, it is capable of achieving a similar LingoQA score to the LingoQA baseline using less than a third of its parameters. Zhous et al. \cite{zhou2024hintspromptenhancingvisual} approach leverages multiple visual and textual inputs to handle understanding: 
The fusion of these inputs is achieved through various methods: Concatenation, where the hints are combined with \gls{clip} tokens; Self-cross-attention, which applies self-attention to \gls{clip} tokens and cross-attention between the output and the hints; Joint cross-attention, where visual tokens act as queries, with both visual tokens and hints serving as keys and values. DriveMM \cite{huang2024drivemmallinonelargemultimodal} is a multimodal model designed to reason across different input modalities, including images and videos. Specifically, it aims to enable a \gls{lvlm} to process and understand \gls{bev} images derived from LiDAR point projections and multi-view videos. The Video Token Sparsification \cite{ma2024videotokensparsificationefficient} model aims to reduce computational effort, by selecting a subset of visual features based on their salience. The salience is determined using a lightweight \gls{cnn} model, called proposal model, which identifies the most relevant features while discarding less informative ones. Finally, Jain et al.\cite{10588373} propose an approach, where object detection and tracking is performed on LiDAR point clouds to retrieve accurate spatial locations for objects in the environment of the ego-vehicle. This information is added to the \gls{vqa} prompt to reduce hallucination, however, their results show that a single \gls{lvlm} cannot process a complex prompt based on the information of the detections.

In addition, some models are trained to solve the End-to-End planning task, but indirectly must also achieve certain understanding of the environment to be able to plan for it. For instance, DriveLM \cite{sima2025drivelmdrivinggraphvisual} proposes a Graph Visual Question Answering pipeline in which a BLIP-2 \gls{vlm} is tasked with answering questions about a scene from perception to prediction to planning to behaviour to motion. SimpleLLM4AD \cite{zheng2024simplellm4adendtoendvisionlanguagemodel} extends the DriveLM baseline by adding a 2D object detection model and passing the bounding boxes, class and color of objects directly into the input of the baseline to extend the information available to the model. LLM-Assist \cite{sharan2023llmassistenhancingclosedloopplanning} combines rule based planning using PDM-closed \cite{dauner2023partingmisconceptionslearningbasedvehicle} with dynamic planning using a \gls{llm} for high uncertainty scenarios. PDM generates 15 trajectories at each planning step. The \gls{llm} is only used when the PDM planner has a high level of uncertainty for the generated trajectories. DriveVLM \cite{tian2024drivevlmconvergenceautonomousdriving} fine-tunes a Qwen-VL 10B model to perform driving scene understanding and planning. It retrieves a driving condition description from the \gls{lvlm} using "Describe the driving condition" and the driving video as input. The obtained driving condition description, the textual description of 2D object detection results and the driving video are then used as input to the same \gls{lvlm} to obtain a scene description focused in the critical objects in the scene and their influence on the ego-vehicle.

\section{Approach}
\label{sec:approach}
\begin{figure}
        \centering
        \includegraphics[width=1\linewidth]{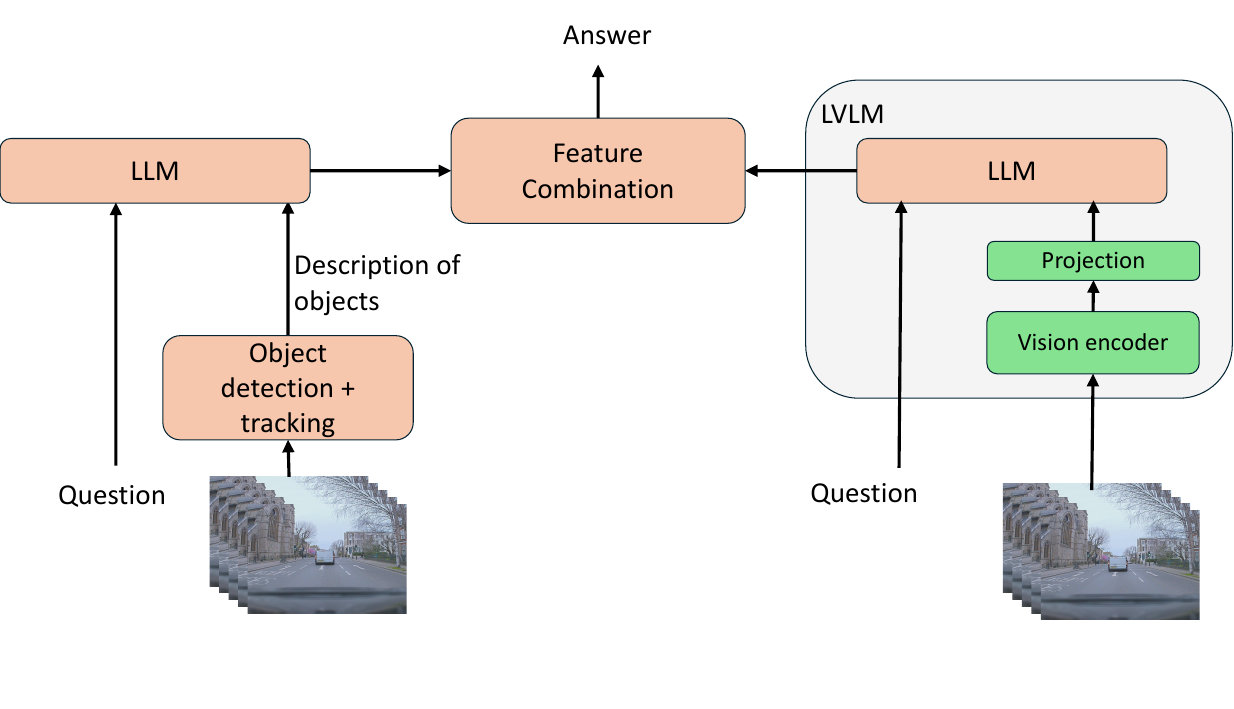}
        \caption{Overview of the approach of this work. At the left side a pre-trained base LLM is getting a textual description of surrounding objects and a question as input. On the right side, a pre-trained LVLM with the same base LLM as on the left-hand side uses the video and the question as input to reason about the visual content of the scene. The information of both models is combined to enhance the overall scene understanding.}
        \label{fig:Overview of the approach}
\end{figure}

A key question is how to integrate the advanced reasoning capabilities of \glspl{llm} with the visual understanding of \glspl{lvlm}. One approach is to combine the textual outputs of the two individual models as input to a third model. This requires three models which leads to increasing computational demands and the data used for processing. Additionally, the output of the models have decreased representational power compared to the internal features of the models. This is a result from the textual output being a mapping from a high dimensional internal feature space to a much smaller discrete output space, allowing to represent one token per position. This process looses a large amount of information about the models internal state.

Another option is to combine the output scores for tokens from both models by merging the score distributions for each output token. This keeps the feature space much richer than relying on the textual output of the models. However, during our experiments it became clear that this is not the ideal way to merge the models as their output distributions differ significantly: \glspl{llm} typically have a broader distribution of scores across different tokens, while \glspl{lvlm} have a more concentrated distribution with higher scores for fewer tokens, leading to a bias towards \gls{lvlm} output in token selection.

In our approach the features per token from both models are merged after individual transformer layers to enable the models to process and modify the combined features. The approach is displayed in \Cref{fig:Overview of the approach}. This enables the models to exchange more detailed information opposed to the approaches presented above. However, it can not be expected that a combination of arbitrarily chosen \glspl{llm} and \glspl{lvlm} perform good results due to their feature space being different. Here, a different feature space between the \gls{llm} and \gls{lvlm} means that similar features from both models correspond to different concepts. Two aspects which ensure model's features to correspond to similar concepts are:

\begin{enumerate}
    \item Models must have similar architectures such that at similar points inside both models, their internal representation corresponds to similar concepts. If this is not the case, features can differ significantly, compromising the feature fusion. Additionally, when the internal dimensionality differs, a mapping mechanism to match both models dimensions needs to be found, which would require additionally tuning. This can be ensured when using a \gls{lvlm} and its base \gls{llm} in the combination as both have the same \gls{llm} architecture. 
    \item Both models should have a similar output space. In the case of \glspl{llm} this corresponds to the models sharing the vocabulary and tokens being mapped to the same index in the output. This is important as it ensures that features close to the output layer correspond to similar concepts, as a result of them being mapped to the same outputs. 
\end{enumerate}
\subsection{Models}
To comply with the forementioned aspects for the feature fusion, the following combinations were tested:
\begin{enumerate}
    \item \gls{llavav} [14B parameter]
    \item \gls{qwen14} [14B parameter]
    \item \gls{qwen35} [3.5B parameter]
\end{enumerate}
\gls{qwen14}, \gls{qwen35} and \gls{llavav} use the Qwen-LLM as base model, fullfillyng the first condition. Moreover, the Qwen-LLM model uses a vocabulary of size 151665, so together with the Qwen-VL they share the same vocabulary and index assignment to tokens. The \gls{llavav} model has a modified vocabulary of size 151647 which uses the same vocabulary and index to token assignment as the Qwen-LLM model. However, it discards the last tokens which are used in Qwen-LLM as special tokens. This might negatively influence the performance when combining \gls{llavav} with Qwen-LLM.

\subsection{3D Understanding module}
\begin{figure}
    \centering
    \includegraphics[width=1\linewidth]{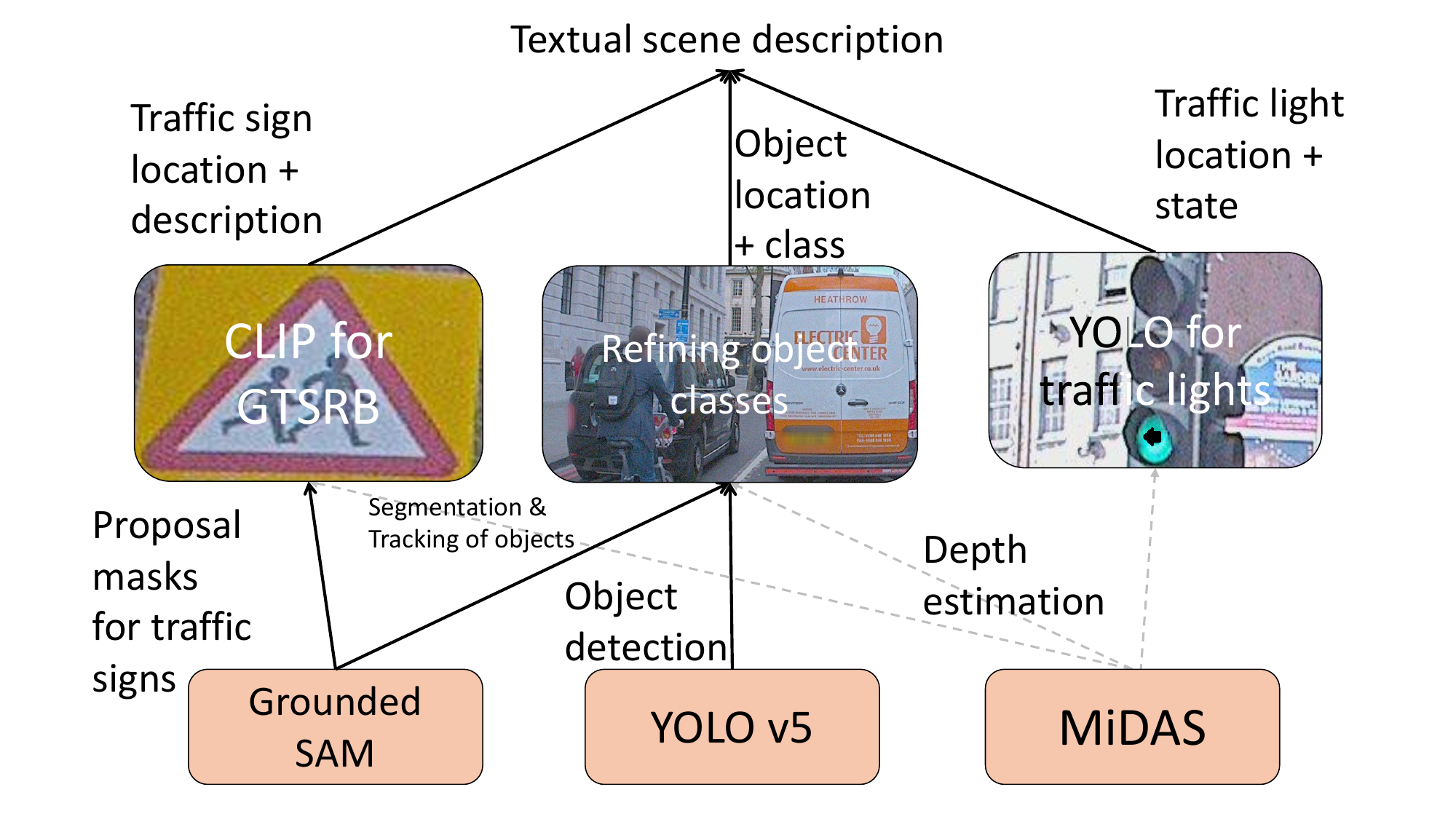}
    \caption{Pipeline for obtaining locations and descriptions for traffic signs, trafic lights and moving objects in the scene.}
    \label{fig:preprocessing}
\end{figure}
The integration of \glspl{llm} and \glspl{lvlm} has been well established. This raises a crucial question: what additional information could enhance the visual processing capabilities of \glspl{lvlm}? For autonomous driving, 3D information may be the key to improve the reasoning capabilities.

Accurate scene understanding is essential for tasks like trajectory planning, which rely on precise 3D localization of traffic participants. Since vehicle interactions occur in 3D space, depth information is critical. However, images inherently lack depth cues, and \glspl{lvlm} struggle with 3D scene comprehension as they are not explicitly provided with this data.

Therefore, we enhance the \gls{lvlm} understanding of the scene through 3D information that is collected through a preprocessing pipeline based on segmentation with GroundedSam \cite{ren2024grounded}, detection with YOLOV5 \cite{glenn_jocher_2020_4154370} monocular depth estimation with MiDas \cite{ranftl2020robustmonoculardepthestimation}. The main structure is presented on \Cref{fig:preprocessing}.

\paragraph{Object detection and tracking}
We begin by focusing on traffic participants, detecting and tracking objects in 2D using Grounded SAM \cite{ren2024grounded}. To enhance segmentation, we integrate a YOLOv5 detector for specific categories of interest. In this work, key objects that significantly influence the ego-vehicle’s driving behavior include trucks, motorbikes, cars, pedestrians, bicycles, traffic lights, and traffic signs.

Nonetheless, extracting 3D information from a monocular image remains essential. To achieve this, we employ the MiDAS depth estimator \cite{ranftl2020robustmonoculardepthestimation}. Depth is assigned to each object by extracting the depth value from the middle pixel of the detection. This results in a final output consisting of a list of detected objects, each labeled with its class and 3D localization relative to the ego-vehicle. A more detailed explanation of this module is provided in the Appendix \Cref{sec:3d_detection}.

Apart from traffic participants, we also notice that \glspl{lvlm} can understand the presence of traffic lights and signs, but have issues extracting the exact meaning of some signs specific to the context and the exact state of the traffic light. To handle this, we propose a module to preprocess this information using a CLIP model and appending the output directly to our textual description of the surrounding objects. The description of those modules is shown in the appendix \Cref{sec:trafficlights}.

\subsection{Merging parameters}
\label{subsec:parameters}
The approach is to combine a \gls{llm} and a \gls{lvlm} to obtain the reasoning capabilities on long language inputs of the \gls{llm} to support the visual reasoning of the \gls{lvlm}. The features of both models will be added together at different stages inside the models and with different weightings. The summing approach keeps the size of the features constant which enables the results to being directly usable in the following model stages. The input to both models differs and therefore, typically does not have the same length, which makes it not possible to combine tokens at the same position. Therefore, only the features of the last produced token will be considered for combination with the feature tokens of the other model. The combination can either be applied pairwise between both models last token or between one models last token and all of the tokens of the other model. This might be interpreted as adding the last tokens information of one model to all the tokens information of the other model, resulting in more substantial modification of the other models features.
The following components contribution to the performance of the combined model will be investigated: 

\paragraph{Head Weights}
The classification heads of the \gls{lvlm} and the \gls{llm} have different weights. They are each optimized to operate correctly on the features that they expect in their input as a result of their individual training. When the features of both models are merged before the classification head, the input to the classification head is modified compared to the input observed during training. This is typically referred to as a change in the feature distribution. To investigate how the weights of the head should idealy be adjusted to be able to deal with the changed input features, differently weighted sums of the head weights are considered. The different configurations are displayed in \Cref{tab:head_weights_definition}. The mathematical formulation of obtaining the combined head weight 
$\mathbf{W_{combined}}\in\mathbb{R}^{\text{model\_dimension} \times \text{vocabulary\_size}}$ is displayed in equation \ref{eq:head_weights}. $p^{head}_{\text{LVLM}}$ corresponds to the head weighting towards the \gls{lvlm} and $p^{head}_{\text{LLM}}$ to the head weighting of the \gls{llm}.

\begin{equation}
    \mathbf{W^{head}_{combined}}=p^{head}_{\text{LLM}} \cdot \mathbf{W^{head}_{\text{LLM}} } + p^{head}_{\text{LVLM}} \cdot \mathbf{W^{head}_{\text{LVLM}} }
    \label{eq:head_weights}
\end{equation}

\begin{table}[ht]
\centering
\caption{The tested head weightings for the merging}
\begin{tabular}{cc}
\hline
\thead{LLM Head Weight\\ $p^{head}_{\text{LLM}}$} & \thead{\gls{lvlm} Head Weight \\$p^{head}_{\text{LVLM}}$} \\
\hline
0.1 & 0.9 \\
0.3 & 0.7 \\
0.5 & 0.5 \\
0.7 & 0.3 \\
0.9 & 0.1 \\
\hline
\end{tabular}

    \label{tab:head_weights_definition}
\end{table}

\paragraph{Weight of feature merging}
In our approach, features of the \gls{llm} and the \gls{lvlm} will be merged by weighted summation. The weights determine the level of contribution of the different models towards the combined features. The weighting factors can be interpreted as the amount of information which is used of each model in consecutive steps. This can lead to conclusions about which models features are more important for a strong performance. The different configurations which are considered in this work are displayed in \Cref{tab:feature_weights_definition}.
\begin{table}[ht]
\centering
\caption{Tested feature weights for the merging.}
\begin{tabular}{cc}
\hline
\thead{LLM Feature Weight \\ $p^{i}_{\text{LLM}}$} & \thead{LVLM Feature Weight \\ $p^{i}_{\text{LVLM}}$} \\
\hline
0.1 & 0.9 \\
0.3 & 0.7 \\
0.5 & 0.5 \\
0.7 & 0.3 \\
0.9 & 0.1 \\
\hline
\end{tabular}

    \label{tab:feature_weights_definition}
\end{table}

\paragraph{Layers after which feature merging is applied}
The layer at which features are merged is crucial, as features from different transformer decoder blocks capture varying levels of abstraction. When both the \gls{llm} and the \gls{lvlm} share the same output vocabulary, merging closer to the output tends to produce better alignment between the modalities. In contrast, merging features too early may impede the models’ abilities to extract meaningful representations independently. In our work, features are considered for merging after each transformer decoder block. Using Qwen 2 as the base \gls{llm}, which comprises 28 transformer decoder blocks, we evaluate three different configurations: merging the features from layers 20 to 28, from layers 25 to 28, or solely from layer 28. This analysis allows us to assess the impact of merging at various depths on overall model performance.

\begin{figure}
    \centering
    \includegraphics[width=1\linewidth]{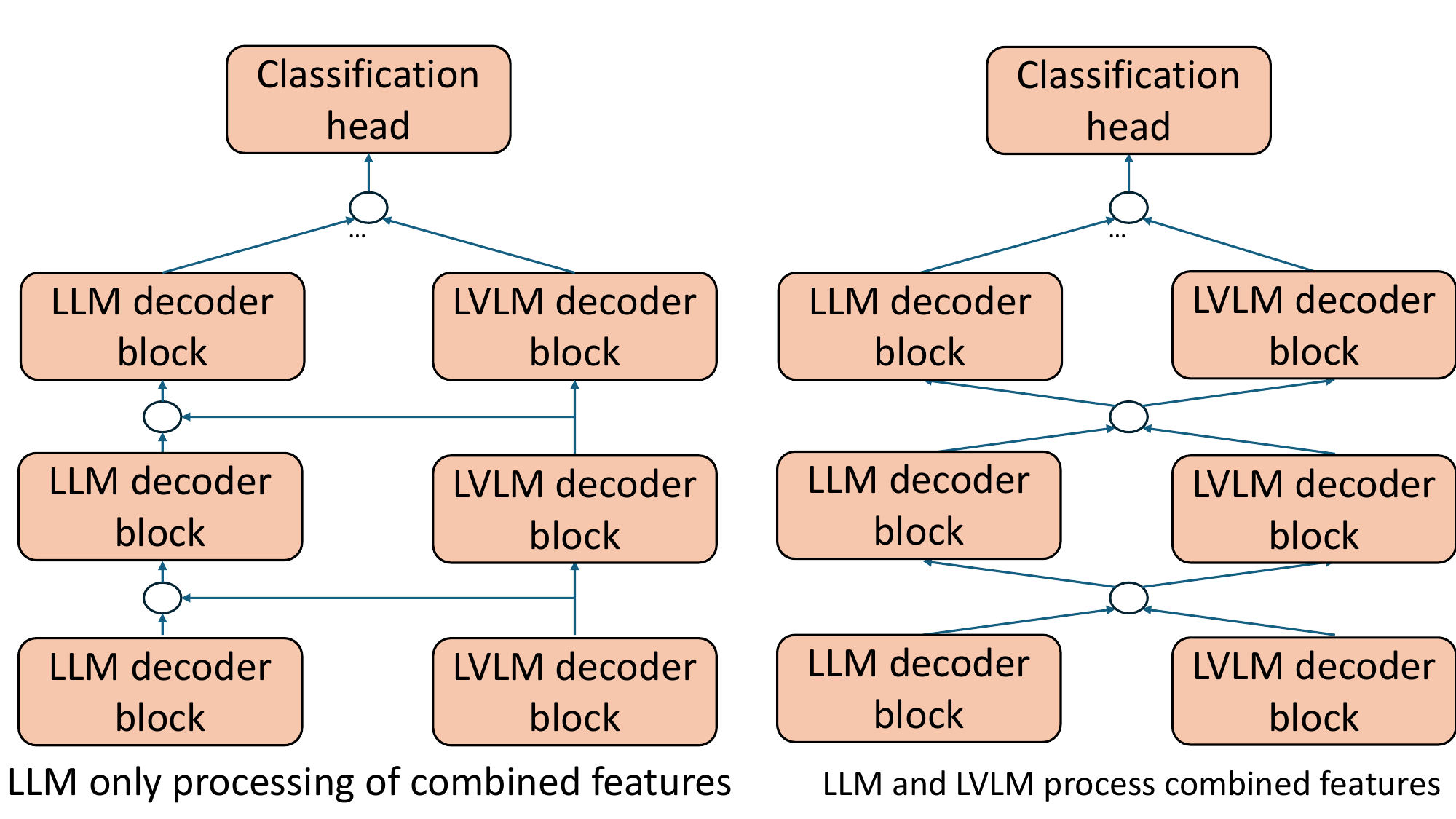}
    \caption{Two different strategies to process the combined features. On the left, the LVLM receives only its own features from the previous layers (False). The LLM processes the combined input of both models. On the right, both models receive the combined features of the previous layer as input (True).}
    \label{fig:feature_combination}
\end{figure}
\paragraph{Branches processing combined data}
The independent branches of the \gls{llm} and the \gls{lvlm} initially process their features separately, enabling each model to extract modality-specific information aligned with its original training objectives. Replacing these distinct features with a combined representation may dilute or obscure some of this individual information. Consequently, it may be advantageous to merge the features into only one branch while allowing the other branch to continue processing its modality-specific information. In our work, we hypothesize that the \gls{llm}, given its superior capability to handle long input sequences, is better suited for reasoning about the more complex, merged features. Thus, we evaluate whether it is beneficial for the \gls{llm} to exclusively process the combined features, while the \gls{lvlm} processes its visual information independently and later contributes to the combined representation that feeds into the \gls{llm}. The branch merging strategy can be seen in \Cref{fig:feature_combination}.

\paragraph{Sum features with weight 1.0 before last layer}
In addition to the feature weights displayed above, it might be beneficial to sum the entire features of the two models. For the QwenLLM architecture,  the normalization of feature vectors is applied in each layer after the skip connection, which leads to different magnitudes of features that are summed to the output of the self-attention compared to weights of [0.5,0.5].

We avoided adding a weight of [1.0,1.0] to the list of feature weight configurations because experiments showed it to be beneficial to weight the features after layer 28 according to the configurations in \Cref{tab:feature_weights_definition} while applying the weights of [1.0,1.0] to those before that layer. Therefore, an additional binary parameter \textit{Sum All} is added.

\subsection{Prompt design}

The input prompts for the \gls{lvlm} and the \gls{llm} are different. As mentioned above, \glspl{lvlm} are not able to reason correctly about long and complicated textual inputs. Therefore, the prompt containing the objects in the surrounding of the ego-vehicle are only being used as input to the \gls{llm}. On the one side. \gls{lvlm} receives for each task only the question and corresponding video data. On the other side, the \gls{llm} receives a prompt composed of three blocks: First, the description and structure of the object data and the general task. Second, the question for the specific task. Lastly, the prompt resulting from the concatenation of the textual information for each object in the scene.
in the appendix \Cref{fig:prompt} the design of a prompt for a sample scene is shown.

\subsection{Experiment}
We compare the different methods of combining the \gls{llm} and \gls{lvlm} to each other on a subset of the LingoQA evaluation set with 200 samples. Consequently, the most promising combinations will be then evaluated in the complete dataset. Inference is performed on a single H100 GPU. In the initial experiments, a random seed of 42 is set and next tokens will be selected based on maximum score, to ensure that the experiments are reproducible.

As the approach is tested with the LingoQA dataset, their proposed metric, the Lingo Judge \cite{marcu2024lingoqavisualquestionanswering} is used as main evaluation. This metric is the output of a DeBERTa-V3 language model fine-tuned using LoRA \cite{hu2021loralowrankadaptationlarge} to assign a score to the quality of model answers on \gls{vqa} tasks on the LingoQA dataset. 
For that, it uses the question, the reference answer and the models answer to output a score between 0 and 1 with 1 being a perfect answer. The LingoQA dataset provides for each question two reference answers and the maximum of the two LingoQA-Judge scores determines the score per question. 
In addition, the ROUGE metric is also presented, to have a result which does not depend on a extra \gls{llm}. However, their focus on recall and exact matches, respectively, could not represent correctly the performance of the models on their tasks, where several answers and formulations can be correct at the same time.

\section{Results}
\label{sec:results}
\begin{table}[t]
    \centering
    \caption{Results of isolated models with \gls{llm} and \gls{lvlm} prompts.}
    \begin{tabular}{ccc}
        \hline
        & \textbf{LLM Prompt} & \textbf{LVLM Prompt} \\ 
        \hline
        \textbf{LLaVA-VL 7B} & $\mathbf{0.419}$ & 0.408 \\ 
        \textbf{Qwen-VL 7B} & $\mathbf{0.419}$ & 0.399 \\ 
        \textbf{Qwen-LLM 7B} & 0.385 & 0.286 \\ 
        \hline
    \end{tabular}
    
    \label{tab:llm_lvml}
\end{table}

As a baseline, we present the results of using only the prepared \gls{llm} or \gls{lvlm} prompt for the isolated models, as seen in \Cref{tab:llm_lvml}. For all the models, the more complete \gls{llm} prompt provides the best scores, suggesting that the \gls{lvlm} models can benefit also from longer and more complex prompts. For the Qwen-LLM the case is similar, but the performance is lower than for the Visual models, indicating that the visual information is still the key for the general scene understanding. In the following paragraphs, the results for the \gls{lvlm}+\gls{llm} combination approach are presented.
\begin{table*}[]
    \centering
    \caption{The best Lingo Scores for each tested model with their respective configuration evaluated on 20\% of the LingoQA dataset.}
\begin{tabular}{llllllrr}
\toprule
Model Name & \thead{Isolate \\ \gls{lvlm} Features} & \thead{Head \\ Weights} & \thead{Feature \\ Weights} & \thead{Merge \\ Layers} & \thead{Sum \\ Weight} & Rouge & \thead{Lingo \\ Mean} \\
\midrule
\gls{llavav} & False & [0.5, 0.5] & [0.9, 0.1] & [25, 26, 27, -1] & True & 0.66 & 0.55 \\
\gls{qwen14} & True & [0.9, 0.1] & [0.9, 0.1] & [25, 26, 27, -1] & True & 0.68 & 0.60 \\
\gls{qwen35} & True & [0.1, 0.9] & [0.3, 0.7] & [25, 26, 27, -1] & False & 0.56 & 0.49 \\
\bottomrule
\label{tab:main results}
\vspace{-0.5cm}
\end{tabular}
\end{table*}

The best results for the three proposed models on the selected subset of the Lingo Dataset are presented in \Cref{tab:main results}. The pure \gls{qwen14} combination achieved a 60\% Lingo Score, marking a nearly 5\% improvement over \gls{llavav}. This difference may be attributed to the modified vocabulary used in LlaVA, as discussed in \Cref{sec:approach}. Meanwhile, \gls{qwen35} achieved a 49\% score. Notably, despite using only 25\% of the parameters, it performed just 10\% below the larger Qwen model. This result highlights the efficiency of our \gls{lvlm}+\gls{llm} combination in processing information while maintaining strong comprehension performance.

Next, in \Cref{table:model_scores_overview_1} we present the results of evaluating the three architectures with the best configuration parameters in the complete Lingo Evaluation dataset, together with the current state of the art (SotA) and other models in the literature. Our approach Lingo score is 0.56. It is worth mentioning that all the models in the literature were fine tuned with the training dataset of the Lingo benchmark. We attempted to improve the performance of a given architecture without the possibility of retraining the model, to see how well was the performance of an approach which cannot have access of labeled data of the target domain source. Furthermore, comparing with the results of \Cref{tab:llm_lvml} there is an improvement of 15\% with respect to the best isolated zero shot models, showing there is actually a large benefit of using the combined information of both architectures. Moreover, comparing with the Lingo model without fine tuning, we show an improvement of 23\%, which shows that our merging architecture effectively takes advantage of general knowledge to understand specific data distributions.

\begin{table}[ht]
\centering
\caption{Model performance scores on the whole LingoQA Evaluation dataset using the mean of the Lingo-Judge scores.}
\begin{tabular}{lcc}
\hline
Name & Fine-tuned & Score \\
\hline
LLaVA 1.5 & Yes & 0.51 \\
InternVL & Yes & 0.58 \\
LingoQA & Yes & 0.61 \\

Drive-OV & Yes & \textbf{0.70} \\
\hline
LingoQA & No & 0.33 \\
\gls{qwen14} (Ours) & No & 0.56 \\
\gls{llavav} (Ours) & No & 0.50 \\
\gls{qwen35} (Ours)& No & 0.45 \\
\hline
\end{tabular}

\label{table:model_scores_overview_1}
\end{table}

\subsection{Ablation study}

In our search for the optimal parameters, we evaluated all possible combinations presented in \Cref{subsec:parameters}. To do this, we aggregated the results for each specific parameter configuration and computed the mean metrics. This approach allowed us to isolate the effects of each parameter under evaluation.

\begin{table}[]
    \centering
    \caption{Lingo Score in the validation set for different Head weights combinations}
\begin{tabular}{lrrrrr}
\toprule
\thead{Head weights \\ Model} & \thead{[0.1, \\ 0.9]} & \thead{[0.3, \\ 0.7]} & \thead{[0.5, \\ 0.5]} & \thead{[0.7, \\ 0.3]} & \thead{[0.9, \\ 0.1]} \\
\midrule
\gls{llavav} & 0.40 & 0.38 & 0.41 & 0.39 & 0.33 \\
\gls{qwen14} & 0.38 & 0.40 & 0.42 & 0.40 & 0.38 \\
\gls{qwen35} & 0.36 & 0.30 & 0.22 & 0.18 & 0.15 \\
\bottomrule
\end{tabular}
    \label{tab:head_weights}
\end{table}

For the head weights, the results are shown in \Cref{tab:head_weights}. For the bigger models, the best performance is obtained when using a balanced weighting between the \gls{llm}'s and \gls{lvlm}'s head features. Nevertheless, the drop in performance is greater when the \gls{llm} features are weighted above the \gls{lvlm} ones.
On the other side, for the smaller Qwen combination the results indicate that the compressed visual features have a greater impact than the textual ones. In this case, for a heavier weighted \gls{lvlm} features the difference is 21\% with respect to heavier \gls{llm} features.

\begin{table}[]
\centering
\caption{Lingo Score in the validation for different feature weights combinations}
\begin{tabular}{lrrrrr}
\toprule
\thead{Feature weights \\ Model} & \thead{[0.1, \\ 0.9]} & \thead{[0.3, \\ 0.7]} & \thead{[0.5, \\ 0.5]} & \thead{[0.7, \\ 0.3]} & \thead{[0.9, \\ 0.1]} \\
\midrule
\gls{llavav} & 0.36 & 0.40 & 0.41 & 0.39 & 0.36 \\
\gls{qwen14} & 0.38 & 0.40 & 0.41 & 0.40 & 0.38 \\
\gls{qwen35} & 0.33 & 0.24 & 0.24 & 0.22 & 0.22 \\
\bottomrule
\end{tabular}
    \label{tab:feature_weights}
\end{table}
Results for feature weighting are shown in \Cref{tab:feature_weights}. It can be seen that the story is similar as for the head weights: The best performance for larger models is achieved with a balanced weighting of the features. However, in this case the influence of the feature combinations is smaller, with only 6\% between the best and worse case and also decreases simetrically with respect to both \gls{lvlm} and LLM weighting. Similarly to the head weights as well, for \gls{qwen35} a stronger priority for the \gls{lvlm} weights resulted in better performance, which supports the idea that for smaller models the visual features contribute more for the performance. 

\begin{table}[]
    \centering
        \caption{Lingo Score for different sets of layers's output that where fused}
    \begin{tabular}{lrrr}
\toprule
\thead{Layers 2 merge \\ Model} & 28 & [20:28] & [25:28] \\
\midrule
\gls{llavav} & 0.38 & 0.33 & 0.44 \\
\gls{qwen14} & 0.39 & 0.35 & 0.44 \\
\gls{qwen35} & 0.26 & 0.25 & 0.28 \\
\bottomrule
\end{tabular}

    \label{tab:layers}
\end{table}

Regarding the feature layers which should be merged, the results are similar for the three tested models. The best performance was found by combining the last four layers of the \gls{lvlm} and \gls{llm}, followed by combining only the last layer before the classification head, and in last place combining the last eight layers. In general combining too many earlier layers can hurt the performance, because those layers are not abstract enough so that the features are well aligned. For \gls{qwen35}, the difference is not  as pronounced as for the larger models, which could mean that the abstraction capability difference between early and late features is not significant enough to be an issue for the merging.

\begin{table}[]
    \centering
    \caption{Lingo Score for whether only the LLM process the features (False), or both LLM and \gls{lvlm} process them (True)}
\begin{tabular}{lrr}
\toprule
Isolate \gls{lvlm} Features & False & True \\
Model &  &  \\
\midrule
\gls{llavav} & 0.36 & 0.40 \\
\gls{qwen14} & 0.36 & 0.44 \\
\gls{qwen35} & 0.21 & 0.30 \\
\bottomrule
\end{tabular}
    
    \label{tab:vlm_features_first}
\end{table}

In \Cref{tab:vlm_features_first} can be seen that it is better to process the combined features through the LLM (\Cref{fig:feature_combination} left) and input the \gls{lvlm} layers only with the previous visual features (\Cref{fig:feature_combination} right). It is then better to not overload the \gls{lvlm} with more text information than it can handle, with similar behavior for the three tested models.

\begin{table}[]
    \centering
     \caption{Lingo Score for whether all the layers are summed with weight [1,1] but the last one or not}
\begin{tabular}{lrr}
\toprule
Sum all & False & True \\
Model &  &  \\
\midrule
\gls{llavav} & 0.40 & 0.37 \\
\gls{qwen14} & 0.40 & 0.39 \\
\gls{qwen35} & 0.30 & 0.22 \\
\bottomrule
\end{tabular}
    \label{tab:sum_feature}
\end{table}

The last parameter to test, is if summing all the layers with weights [1,1] improves the performance of the models. According to the results in \Cref{tab:sum_feature}, is not the case, but the difference is less than 1\% for the bigger models, which could indicate that it may be also statistical noise in the measurement. For \gls{qwen35}, the difference of 8\% shows a clear benefit of not combining the features with a [1,1] weighting vector.

\section{Discussion}
\label{sec:discussion}

The most notable insight from our results is the apparent inconsistency between the optimal parameters of the best-performing models and the trends observed in the ablation study. For instance, for both head weights and feature weights we found in the ablation studies that a balanced weighting for \gls{llm} and \gls{lvlm} should yield the best results. However, in \Cref{tab:main results}, the top-performing models consistently favor an unbalanced weighting toward the \gls{llm} head. Moreover, in two of these models, the unbalanced setting for feature weights also outperforms the balanced configuration. A similar contradiction is observed for the "Sum All" parameter. We attribute these inconsistencies to cross-correlations between parameters, meaning that the impact of one parameter is influenced by the values of others. Such interdependencies are not captured when analyzing ablation results in isolation, as these results are marginalized over all other settings.

In contrast, for the remaining parameters, the findings are more conclusive. For the "Isolate \gls{lvlm} features" parameter, we demonstrate that it is beneficial to process both \gls{llm} and \gls{lvlm} features solely within the \gls{llm} decoder blocks while isolating the \gls{lvlm} decoder blocks to handle only visual features. This configuration leverages the \gls{llm}’s superior ability to process complex prompts enriched with 3D information, while allowing the \gls{lvlm} to focus on semantic visual understanding derived from camera input, thereby complementing the\gls{llm}’s reasoning with perceptual data. Furthermore, we prove that late fusion of the features enhances the alignment between the \gls{llm} and \gls{lvlm}'s encoder output, improving the synergy between text and visual features. Fusing only the final layer proves insufficient, indicating that the last four layers contain valuable information for the general understanding task.

Although our primary objective was to develop a zero-shot approach that eliminates the need for fine-tuning, future work could explore how retraining on the target data distribution may further enhance performance. Moreover, a deeper investigation into the cross-dependencies of fusion parameters could provide valuable insights into how specific settings influence one another—potentially enabling a more systematic and optimized fusion strategy for combining \glspl{llm} and \glspl{lvlm}.

Finally, while our evaluation focused on the LingoQA dataset, our method is designed without any dataset-specific training, suggesting potential generalization capabilities. Nonetheless, validating this assumption through quantitative evaluation on other VQA datasets would be essential. Unfortunately, as detailed in Appendix \Cref{table:language_datasets}, no other publicly available datasets currently offer the necessary characteristics to allow for a direct comparison.
\section{Conclusion}
\label{sec:Conclusion}
We introduce a novel zero-shot multimodal approach for 3D scene understanding in traffic scenarios. Our method enhances the reasoning capabilities of a \gls{llm} by integrating 3D information distilled from object detectors. This enriched spatial data is fused with visual features extracted from a \gls{lvlm}, enabling the model to simultaneously interpret the semantic content of images and the spatial structure of the 3D environment.

We further examine various feature fusion configurations, providing practical insights into optimizing the interplay between \glspl{llm} and \glspl{lvlm} for improved scene comprehension. To the best of our knowledge, our approach achieves the highest LingoQA score for zero-shot multimodal scene understanding. Importantly, our framework generalizes well and delivers robust performance without the need for fine-tuning on the target domain, making it an adaptable solution for a wide range of real-world applications.

\clearpage

{
    \small
    \bibliographystyle{ieeenat_fullname}
    \bibliography{main}

\begin{thebibliography}{43}
\providecommand{\natexlab}[1]{#1}
\providecommand{\url}[1]{\texttt{#1}}
\expandafter\ifx\csname urlstyle\endcsname\relax
  \providecommand{\doi}[1]{doi: #1}\else
  \providecommand{\doi}{doi: \begingroup \urlstyle{rm}\Url}\fi

\bibitem[Arai et~al.(2024)Arai, Miwa, Sasaki, Yamaguchi, Watanabe, Aoki, and
  Yamamoto]{arai2024covlacomprehensivevisionlanguageactiondataset}
Hidehisa Arai, Keita Miwa, Kento Sasaki, Yu Yamaguchi, Kohei Watanabe, Shunsuke
  Aoki, and Issei Yamamoto.
\newblock Covla: Comprehensive vision-language-action dataset for autonomous
  driving, 2024.

\bibitem[Cao et~al.(2023)Cao, Zhou, Ma, Ye, Cui, Tang, Cao, Liang, Wang, Rehg,
  and Zheng]{tencent2023maplm}
Xu Cao, Tong Zhou, Yunsheng Ma, Wenqian Ye, Can Cui, Kun Tang, Zhipeng Cao,
  Kaizhao Liang, Ziran Wang, James Rehg, and Chao Zheng.
\newblock Maplm: A real-world large-scale vision-language dataset for map and
  traffic scene understanding.
\newblock \url{https://github.com/LLVM-AD/MAPLM}, 2023.

\bibitem[Choudhary et~al.(2023)Choudhary, Dewangan, Chandhok, Priyadarshan,
  Jain, Singh, Srivastava, Jatavallabhula, and
  Krishna]{choudhary2023talk2bevlanguageenhancedbirdseyeview}
Tushar Choudhary, Vikrant Dewangan, Shivam Chandhok, Shubham Priyadarshan,
  Anushka Jain, Arun~K. Singh, Siddharth Srivastava, Krishna~Murthy
  Jatavallabhula, and K.~Madhava Krishna.
\newblock Talk2bev: Language-enhanced bird's-eye view maps for autonomous
  driving, 2023.

\bibitem[Chu et~al.(2024)Chu, Qiao, Zhang, Xu, Wei, Yang, Sun, Hu, Lin, Zhang,
  and Shen]{chu2024mobilevlmv2fasterstronger}
Xiangxiang Chu, Limeng Qiao, Xinyu Zhang, Shuang Xu, Fei Wei, Yang Yang,
  Xiaofei Sun, Yiming Hu, Xinyang Lin, Bo Zhang, and Chunhua Shen.
\newblock Mobilevlm v2: Faster and stronger baseline for vision language model,
  2024.

\bibitem[Dauner et~al.(2023)Dauner, Hallgarten, Geiger, and
  Chitta]{dauner2023partingmisconceptionslearningbasedvehicle}
Daniel Dauner, Marcel Hallgarten, Andreas Geiger, and Kashyap Chitta.
\newblock Parting with misconceptions about learning-based vehicle motion
  planning, 2023.

\bibitem[Deruyttere et~al.(2019)Deruyttere, Vandenhende, Grujicic, Van~Gool,
  and Moens]{Deruyttere_2019}
Thierry Deruyttere, Simon Vandenhende, Dusan Grujicic, Luc Van~Gool, and
  Marie-Francine Moens.
\newblock Talk2car: Taking control of your self-driving car.
\newblock In \emph{Proceedings of the 2019 Conference on Empirical Methods in
  Natural Language Processing and the 9th International Joint Conference on
  Natural Language Processing (EMNLP-IJCNLP)}. Association for Computational
  Linguistics, 2019.

\bibitem[Guo et~al.(2023)Guo, Zhang, Zhu, Tang, Ma, Han, Chen, Gao, Li, Li, and
  Heng]{guo2023pointbindpointllmaligning}
Ziyu Guo, Renrui Zhang, Xiangyang Zhu, Yiwen Tang, Xianzheng Ma, Jiaming Han,
  Kexin Chen, Peng Gao, Xianzhi Li, Hongsheng Li, and Pheng-Ann Heng.
\newblock Point-bind and point-llm: Aligning point cloud with multi-modality
  for 3d understanding, generation, and instruction following, 2023.

\bibitem[Hartsock and Rasool(2024)]{Hartsock_2024}
Iryna Hartsock and Ghulam Rasool.
\newblock Vision-language models for medical report generation and visual
  question answering: a review.
\newblock \emph{Frontiers in Artificial Intelligence}, 7, 2024.

\bibitem[Hu et~al.(2021)Hu, Shen, Wallis, Allen-Zhu, Li, Wang, Wang, and
  Chen]{hu2021loralowrankadaptationlarge}
Edward~J. Hu, Yelong Shen, Phillip Wallis, Zeyuan Allen-Zhu, Yuanzhi Li, Shean
  Wang, Lu Wang, and Weizhu Chen.
\newblock Lora: Low-rank adaptation of large language models, 2021.

\bibitem[Huang et~al.(2024)Huang, Feng, Yan, Xiao, Jie, Zhong, Liang, and
  Ma]{huang2024drivemmallinonelargemultimodal}
Zhijian Huang, Chengjian Feng, Feng Yan, Baihui Xiao, Zequn Jie, Yujie Zhong,
  Xiaodan Liang, and Lin Ma.
\newblock Drivemm: All-in-one large multimodal model for autonomous driving,
  2024.

\bibitem[Inoue et~al.(2023)Inoue, Yada, Tanahashi, and
  Yamaguchi]{inoue2023nuscenesmqaintegratedevaluationcaptions}
Yuichi Inoue, Yuki Yada, Kotaro Tanahashi, and Yu Yamaguchi.
\newblock Nuscenes-mqa: Integrated evaluation of captions and qa for autonomous
  driving datasets using markup annotations, 2023.

\bibitem[Jain et~al.(2024)Jain, Thapa, Chen, Abbott, and Sarkar]{10588373}
Sandesh Jain, Surendrabikram Thapa, Kuan-Ting Chen, A.~Lynn Abbott, and Abhijit
  Sarkar.
\newblock Semantic understanding of traffic scenes with large vision language
  models.
\newblock In \emph{2024 IEEE Intelligent Vehicles Symposium (IV)}, pages
  1580--1587, 2024.

\bibitem[Jocher et~al.(2020)Jocher, Stoken, Borovec, NanoCode012,
  ChristopherSTAN, Changyu, Laughing, tkianai, Hogan, lorenzomammana, yxNONG,
  AlexWang1900, Diaconu, Marc, wanghaoyang0106, ml5ah, Doug, Ingham, Frederik,
  Guilhen, Hatovix, Poznanski, Fang, Yu, changyu98, Wang, Gupta, Akhtar,
  PetrDvoracek, and Rai]{glenn_jocher_2020_4154370}
Glenn Jocher, Alex Stoken, Jirka Borovec, NanoCode012, ChristopherSTAN, Liu
  Changyu, Laughing, tkianai, Adam Hogan, lorenzomammana, yxNONG, AlexWang1900,
  Laurentiu Diaconu, Marc, wanghaoyang0106, ml5ah, Doug, Francisco Ingham,
  Frederik, Guilhen, Hatovix, Jake Poznanski, Jiacong Fang, Lijun Yu,
  changyu98, Mingyu Wang, Naman Gupta, Osama Akhtar, PetrDvoracek, and Prashant
  Rai.
\newblock ultralytics/yolov5: v3.1 - bug fixes and performance improvements,
  2020.

\bibitem[Kim et~al.(2018)Kim, Rohrbach, Darrell, Canny, and
  Akata]{kim2018textualexplanationsselfdrivingvehicles}
Jinkyu Kim, Anna Rohrbach, Trevor Darrell, John Canny, and Zeynep Akata.
\newblock Textual explanations for self-driving vehicles, 2018.

\bibitem[Liu et~al.(2023)Liu, Li, Wu, and Lee]{liu2023visualinstructiontuning}
Haotian Liu, Chunyuan Li, Qingyang Wu, and Yong~Jae Lee.
\newblock Visual instruction tuning, 2023.

\bibitem[Lu et~al.(2024)Lu, Yao, Tu, Shao, Ma, and
  Zhu]{lu2024lvlmsobtaindriverslicense}
Yuhang Lu, Yichen Yao, Jiadong Tu, Jiangnan Shao, Yuexin Ma, and Xinge Zhu.
\newblock Can lvlms obtain a driver's license? a benchmark towards reliable agi
  for autonomous driving, 2024.

\bibitem[Ma et~al.(2024)Ma, Abdelraouf, Gupta, Wang, and
  Han]{ma2024videotokensparsificationefficient}
Yunsheng Ma, Amr Abdelraouf, Rohit Gupta, Ziran Wang, and Kyungtae Han.
\newblock Video token sparsification for efficient multimodal llms in
  autonomous driving, 2024.

\bibitem[Malla et~al.(2022)Malla, Choi, Dwivedi, Choi, and
  Li]{malla2022dramajointrisklocalization}
Srikanth Malla, Chiho Choi, Isht Dwivedi, Joon~Hee Choi, and Jiachen Li.
\newblock Drama: Joint risk localization and captioning in driving, 2022.

\bibitem[Marcu et~al.(2024)Marcu, Chen, Hünermann, Karnsund, Hanotte,
  Chidananda, Nair, Badrinarayanan, Kendall, Shotton, Arani, and
  Sinavski]{marcu2024lingoqavisualquestionanswering}
Ana-Maria Marcu, Long Chen, Jan Hünermann, Alice Karnsund, Benoit Hanotte,
  Prajwal Chidananda, Saurabh Nair, Vijay Badrinarayanan, Alex Kendall, Jamie
  Shotton, Elahe Arani, and Oleg Sinavski.
\newblock Lingoqa: Visual question answering for autonomous driving, 2024.

\bibitem[Moor et~al.(2023)Moor, Huang, Wu, Yasunaga, Zakka, Dalmia, Reis,
  Rajpurkar, and Leskovec]{moor2023medflamingomultimodalmedicalfewshot}
Michael Moor, Qian Huang, Shirley Wu, Michihiro Yasunaga, Cyril Zakka, Yash
  Dalmia, Eduardo~Pontes Reis, Pranav Rajpurkar, and Jure Leskovec.
\newblock Med-flamingo: a multimodal medical few-shot learner, 2023.

\bibitem[Nie et~al.(2024)Nie, Peng, Wang, Cai, Han, Xu, and
  Zhang]{nie2024reason2driveinterpretablechainbasedreasoning}
Ming Nie, Renyuan Peng, Chunwei Wang, Xinyue Cai, Jianhua Han, Hang Xu, and Li
  Zhang.
\newblock Reason2drive: Towards interpretable and chain-based reasoning for
  autonomous driving, 2024.

\bibitem[OpenAI et~al.(2024)OpenAI, Achiam, Adler, Agarwal, Ahmad, and
  and]{openai2024gpt4technicalreport}
OpenAI, Josh Achiam, Steven Adler, Sandhini Agarwal, Lama Ahmad, and
  Ilge~Akkaya and.
\newblock Gpt-4 technical report, 2024.

\bibitem[Park et~al.(2024)Park, Lee, Kang, Choi, Park, Cho, Lee, and
  Kim]{10495690}
SungYeon Park, MinJae Lee, JiHyuk Kang, Hahyeon Choi, Yoonah Park, Juhwan Cho,
  Adam Lee, and DongKyu Kim.
\newblock Vlaad: Vision and language assistant for autonomous driving.
\newblock In \emph{2024 IEEE/CVF Winter Conference on Applications of Computer
  Vision Workshops (WACVW)}, pages 980--987, 2024.

\bibitem[Qian et~al.(2024)Qian, Chen, Zhuo, Jiao, and
  Jiang]{qian2024nuscenesqamultimodalvisualquestion}
Tianwen Qian, Jingjing Chen, Linhai Zhuo, Yang Jiao, and Yu-Gang Jiang.
\newblock Nuscenes-qa: A multi-modal visual question answering benchmark for
  autonomous driving scenario, 2024.

\bibitem[Ranftl et~al.(2020)Ranftl, Lasinger, Hafner, Schindler, and
  Koltun]{ranftl2020robustmonoculardepthestimation}
René Ranftl, Katrin Lasinger, David Hafner, Konrad Schindler, and Vladlen
  Koltun.
\newblock Towards robust monocular depth estimation: Mixing datasets for
  zero-shot cross-dataset transfer, 2020.

\bibitem[Ren et~al.(2024)Ren, Liu, Zeng, Lin, Li, Cao, Chen, Huang, Chen, Yan,
  Zeng, Zhang, Li, Yang, Li, Jiang, and Zhang]{ren2024grounded}
Tianhe Ren, Shilong Liu, Ailing Zeng, Jing Lin, Kunchang Li, He Cao, Jiayu
  Chen, Xinyu Huang, Yukang Chen, Feng Yan, Zhaoyang Zeng, Hao Zhang, Feng Li,
  Jie Yang, Hongyang Li, Qing Jiang, and Lei Zhang.
\newblock Grounded sam: Assembling open-world models for diverse visual tasks,
  2024.

\bibitem[Sachdeva et~al.(2023)Sachdeva, Agarwal, Chundi, Roelofs, Li,
  Kochenderfer, Choi, and
  Dariush]{sachdeva2023rank2tellmultimodaldrivingdataset}
Enna Sachdeva, Nakul Agarwal, Suhas Chundi, Sean Roelofs, Jiachen Li, Mykel
  Kochenderfer, Chiho Choi, and Behzad Dariush.
\newblock Rank2tell: A multimodal driving dataset for joint importance ranking
  and reasoning, 2023.

\bibitem[Shao et~al.(2023)Shao, Hu, Wang, Waslander, Liu, and
  Li]{shao2023lmdriveclosedloopendtoenddriving}
Hao Shao, Yuxuan Hu, Letian Wang, Steven~L. Waslander, Yu Liu, and Hongsheng
  Li.
\newblock Lmdrive: Closed-loop end-to-end driving with large language models,
  2023.

\bibitem[Sharan et~al.(2023)Sharan, Pittaluga, G, and
  Chandraker]{sharan2023llmassistenhancingclosedloopplanning}
S~P Sharan, Francesco Pittaluga, Vijay Kumar~B G, and Manmohan Chandraker.
\newblock Llm-assist: Enhancing closed-loop planning with language-based
  reasoning, 2023.

\bibitem[Sima et~al.(2025)Sima, Renz, Chitta, Chen, Zhang, Xie, Beißwenger,
  Luo, Geiger, and Li]{sima2025drivelmdrivinggraphvisual}
Chonghao Sima, Katrin Renz, Kashyap Chitta, Li Chen, Hanxue Zhang, Chengen Xie,
  Jens Beißwenger, Ping Luo, Andreas Geiger, and Hongyang Li.
\newblock Drivelm: Driving with graph visual question answering, 2025.

\bibitem[Stallkamp et~al.(2011)Stallkamp, Schlipsing, Salmen, and
  Igel]{6033395}
Johannes Stallkamp, Marc Schlipsing, Jan Salmen, and Christian Igel.
\newblock The german traffic sign recognition benchmark: A multi-class
  classification competition.
\newblock In \emph{The 2011 International Joint Conference on Neural Networks},
  pages 1453--1460, 2011.

\bibitem[{SungYeon Park} et~al.(){SungYeon Park}, {MinJae Lee}, {JiHyuk Kang},
  {Hahyeon Choi}, {Yoonah Park}, {Juhwan Cho}, {Adam Lee}, and {DongKyu
  Kim}]{SungYeonPark.}
{SungYeon Park}, {MinJae Lee}, {JiHyuk Kang}, {Hahyeon Choi}, {Yoonah Park},
  {Juhwan Cho}, {Adam Lee}, and {DongKyu Kim}.
\newblock Vlaad: Vision and language assistant for autonomous driving.

\bibitem[Tian et~al.(2024)Tian, Gu, Li, Liu, Wang, Zhao, Zhan, Jia, Lang, and
  Zhao]{tian2024drivevlmconvergenceautonomousdriving}
Xiaoyu Tian, Junru Gu, Bailin Li, Yicheng Liu, Yang Wang, Zhiyong Zhao, Kun
  Zhan, Peng Jia, Xianpeng Lang, and Hang Zhao.
\newblock Drivevlm: The convergence of autonomous driving and large
  vision-language models, 2024.

\bibitem[Wang et~al.(2024)Wang, Yu, Jiang, Lan, Shi, Chang, Kautz, Li, and
  Alvarez]{wang2024omnidriveholisticllmagentframework}
Shihao Wang, Zhiding Yu, Xiaohui Jiang, Shiyi Lan, Min Shi, Nadine Chang, Jan
  Kautz, Ying Li, and Jose~M. Alvarez.
\newblock Omnidrive: A holistic llm-agent framework for autonomous driving with
  3d perception, reasoning and planning, 2024.

\bibitem[Wu et~al.(2023{\natexlab{a}})Wu, Han, Wang, Dong, Zhang, and
  Shen]{wu2023referringmultiobjecttracking}
Dongming Wu, Wencheng Han, Tiancai Wang, Xingping Dong, Xiangyu Zhang, and
  Jianbing Shen.
\newblock Referring multi-object tracking, 2023{\natexlab{a}}.

\bibitem[Wu et~al.(2023{\natexlab{b}})Wu, Han, Wang, Liu, Zhang, and
  Shen]{wu2023languagepromptautonomousdriving}
Dongming Wu, Wencheng Han, Tiancai Wang, Yingfei Liu, Xiangyu Zhang, and
  Jianbing Shen.
\newblock Language prompt for autonomous driving, 2023{\natexlab{b}}.

\bibitem[Xu et~al.(2021)Xu, Huang, and
  Liu]{xu2021sutdtrafficqaquestionansweringbenchmark}
Li Xu, He Huang, and Jun Liu.
\newblock Sutd-trafficqa: A question answering benchmark and an efficient
  network for video reasoning over traffic events, 2021.

\bibitem[Xu et~al.()Xu, Wang, Wang, Chen, Pang, and Lin]{Xu.31.08.2023}
Runsen Xu, Xiaolong Wang, Tai Wang, Yilun Chen, Jiangmiao Pang, and Dahua Lin.
\newblock Pointllm: Empowering large language models to understand point
  clouds.

\bibitem[Zhang et~al.(2024)Zhang, Huang, Gao, Chen, and
  Lv]{zhang2024wiseadknowledgeaugmentedendtoend}
Songyan Zhang, Wenhui Huang, Zihui Gao, Hao Chen, and Chen Lv.
\newblock Wisead: Knowledge augmented end-to-end autonomous driving with
  vision-language model, 2024.

\bibitem[Zhang et~al.(2025)Zhang, Xu, Usuyama, Xu, Bagga, Tinn, Preston, Rao,
  Wei, Valluri, Wong, Tupini, Wang, Mazzola, Shukla, Liden, Gao, Crabtree,
  Piening, Bifulco, Lungren, Naumann, Wang, and
  Poon]{zhang2025biomedclipmultimodalbiomedicalfoundation}
Sheng Zhang, Yanbo Xu, Naoto Usuyama, Hanwen Xu, Jaspreet Bagga, Robert Tinn,
  Sam Preston, Rajesh Rao, Mu Wei, Naveen Valluri, Cliff Wong, Andrea Tupini,
  Yu Wang, Matt Mazzola, Swadheen Shukla, Lars Liden, Jianfeng Gao, Angela
  Crabtree, Brian Piening, Carlo Bifulco, Matthew~P. Lungren, Tristan Naumann,
  Sheng Wang, and Hoifung Poon.
\newblock Biomedclip: a multimodal biomedical foundation model pretrained from
  fifteen million scientific image-text pairs, 2025.

\bibitem[Zheng et~al.(2024)Zheng, Zhao, Gong, Zhu, and
  Wu]{zheng2024simplellm4adendtoendvisionlanguagemodel}
Peiru Zheng, Yun Zhao, Zhan Gong, Hong Zhu, and Shaohua Wu.
\newblock Simplellm4ad: An end-to-end vision-language model with graph visual
  question answering for autonomous driving, 2024.

\bibitem[Zhou et~al.(2024)Zhou, Gao, Ye, Chen, Chen, Cao, and
  Qi]{zhou2024hintspromptenhancingvisual}
Hao Zhou, Zhanning Gao, Maosheng Ye, Zhili Chen, Qifeng Chen, Tongyi Cao, and
  Honggang Qi.
\newblock Hints of prompt: Enhancing visual representation for multimodal llms
  in autonomous driving, 2024.

\bibitem[Zhou et~al.(2025)Zhou, Larintzakis, Guo, Zimmer, Liu, Cao, Zhang,
  Lakshminarasimhan, Strand, and
  Knoll]{zhou2025tumtrafficvideoqabenchmarkunifiedspatiotemporal}
Xingcheng Zhou, Konstantinos Larintzakis, Hao Guo, Walter Zimmer, Mingyu Liu,
  Hu Cao, Jiajie Zhang, Venkatnarayanan Lakshminarasimhan, Leah Strand, and
  Alois~C. Knoll.
\newblock Tumtraffic-videoqa: A benchmark for unified spatio-temporal video
  understanding in traffic scenes, 2025.

\end{thebibliography}
}

\clearpage
\setcounter{page}{1}
\maketitlesupplementary

\section{VQA Autonomous Driving Datasets}
In \Cref{table:language_datasets} we present the datasets known for VQA in autonomous driving. Only LingoQA fullfills the necessary conditions for this study: Video modality, video-level annotation and publicly available. 
\begin{table*}
\begin{tabular}{l|l|l|l|l|l}
\hline
\multirow{2}{*}{\textbf{Dataset Name}} & \multirow{2}{*}{\textbf{Modalities}} & \multirow{2}{*}{\textbf{Base dataset}} & \multirow{2}{*}{\textbf{QA}} & 
\textbf{Video-level} & \multirow{2}{*}{\textbf{Annotation description}} \\ 
& & & & \textbf{annotation} &\\

\hline
BDD-X\cite{kim2018textualexplanationsselfdrivingvehicles} & video & BDD & no & yes & Action + reasoning \\
Talk2Car\cite{Deruyttere_2019} & video, point cloud & nuScenes & no & - & Instruction for vehicle to execute \\
SUTD-TrafficQA\cite{xu2021sutdtrafficqaquestionansweringbenchmark} * & video & & yes & yes & Driving QA, not from ego perspective \\
DRAMA\cite{malla2022dramajointrisklocalization} * & video & & no & yes & Driving scene captioning \\
nuScenes-QA\cite{qian2024nuscenesqamultimodalvisualquestion} & video, point cloud & nuScenes & yes & no & Driving QA \\
NuPrompt\cite{wu2023languagepromptautonomousdriving} & video, point cloud & nuScenes & no & yes & Object detection \\
DriveLM\cite{sima2025drivelmdrivinggraphvisual} & video, point cloud & nuScenes & yes & no & Driving QA image, Caption Video \\
Rank2Tell\cite{sachdeva2023rank2tellmultimodaldrivingdataset} & video, point cloud & & no & yes & Object importance + Reasoning \\
MAPLM-QA\cite{tencent2023maplm} & image, point cloud & & yes & no & Driving QA \\
LINGO-QA\cite{marcu2024lingoqavisualquestionanswering} & video & & yes & yes & Driving QA \\
lmdrive\cite{shao2023lmdriveclosedloopendtoenddriving} & video, point cloud & carla-based & no & - & Level of throttle + turning angle \\
vlaad\cite{10495690} & video & bdd & yes & yes & Driving QA \\
vlaad\cite{10495690} & video & hadhri & yes & yes & Driving QA \\
talk2bev\cite{choudhary2023talk2bevlanguageenhancedbirdseyeview} * & video, point cloud & nuScenes & no & yes & Central scene object using LVLMs\\
refer-kitti\cite{wu2023referringmultiobjecttracking} & video, point cloud & kitti & no & no & Object referral \\
reason2drive\cite{nie2024reason2driveinterpretablechainbasedreasoning} * & video, point cloud & nuScenes & yes & yes & Object referral + Driving QA \\
nuScenes-MQA\cite{inoue2023nuscenesmqaintegratedevaluationcaptions} & video, point cloud & nuScenes & no & no & Object referral \\
IDKB\cite{lu2024lvlmsobtaindriverslicense} & image &  & yes & no & Driving QA \\
CoVLA\cite{arai2024covlacomprehensivevisionlanguageactiondataset} & image, point cloud & & no & yes & Driving scene captioning \\
OmniDrive\cite{wang2024omnidriveholisticllmagentframework} & image, point cloud & nuScenes & yes & no & Trajectory QA \\
TUMTraffic-VideoQA\cite{zhou2025tumtrafficvideoqabenchmarkunifiedspatiotemporal} & image &  & yes & yes & Driving QA not from ego perspective \\
\hline
\end{tabular}
\caption{Overview of datasets providing language data for driving. Datasets indicated by * are not publicly or only partially available. We contacted the authors but did not get an answer}
\label{table:language_datasets}
\end{table*}

\section{3D Detection and Tracking}
\label{sec:3d_detection}
The first step is to detect and track objects of interest using Grounded SAM \cite{ren2024grounded}. This model is capable of detecting and tracking arbitrary objects described by text. In this work, all objects in the vicinity of the ego-vehicle are regarded as of interest if they belong to a given class. Tracking is crucial for the \gls{llm}, as it enables the analysis of object positions across frames which can lead to detecting movement patterns, a task which is difficult for the \gls{llm} to perform without instance-specific information. To achieve this, each object is assigned a cross-frame ID. Segmentation provides pixel-wise assignment to instances. 

It is asummed that only monocular camera information is available. Consequently, depth estimation is performed for each pixel in a frame using MiDaS \cite{ranftl2020robustmonoculardepthestimation}. However, depth values are relative to other pixels in the same image, which complicates cross-frame comparisons. To address this, depth normalization is applied using reference points known to maintain consistent distances across frames, with the distance between two points on the hood of the ego-vehicle used as the normalization metric. Additionally, shadows can cause errors in depth estimation, and road points near the ego-vehicle are often obscured by them. Despite these challenges, this method provides sufficient relative approximations of distance to analyze the changes in object positions over time. 

The segmentation process yields depth values for each pixel of an object, reducing the noise introduced by background depth estimations. To streamline the input, only the mean distance for each object is included. Grounded SAM is outperformed by YOLO 5 \cite{glenn_jocher_2020_4154370}  when classifying objects. Therefore, additionally YOLO 5 detects objects of categories of interest. Between each of YOLOs detection and those of Grounded SAM, the intersection over union is calculated. If the value is above 0.35, the detected object by Grounded SAM is assigned the class determined by YOLO. If an object detected by YOLO is not detected by Grounded SAM, it is added using the tracking ID -1. 

\section{Traffic light detection and state recognition}
\label{sec:trafficlights}
\begin{figure}
    \centering
    \includegraphics[width=1\linewidth]{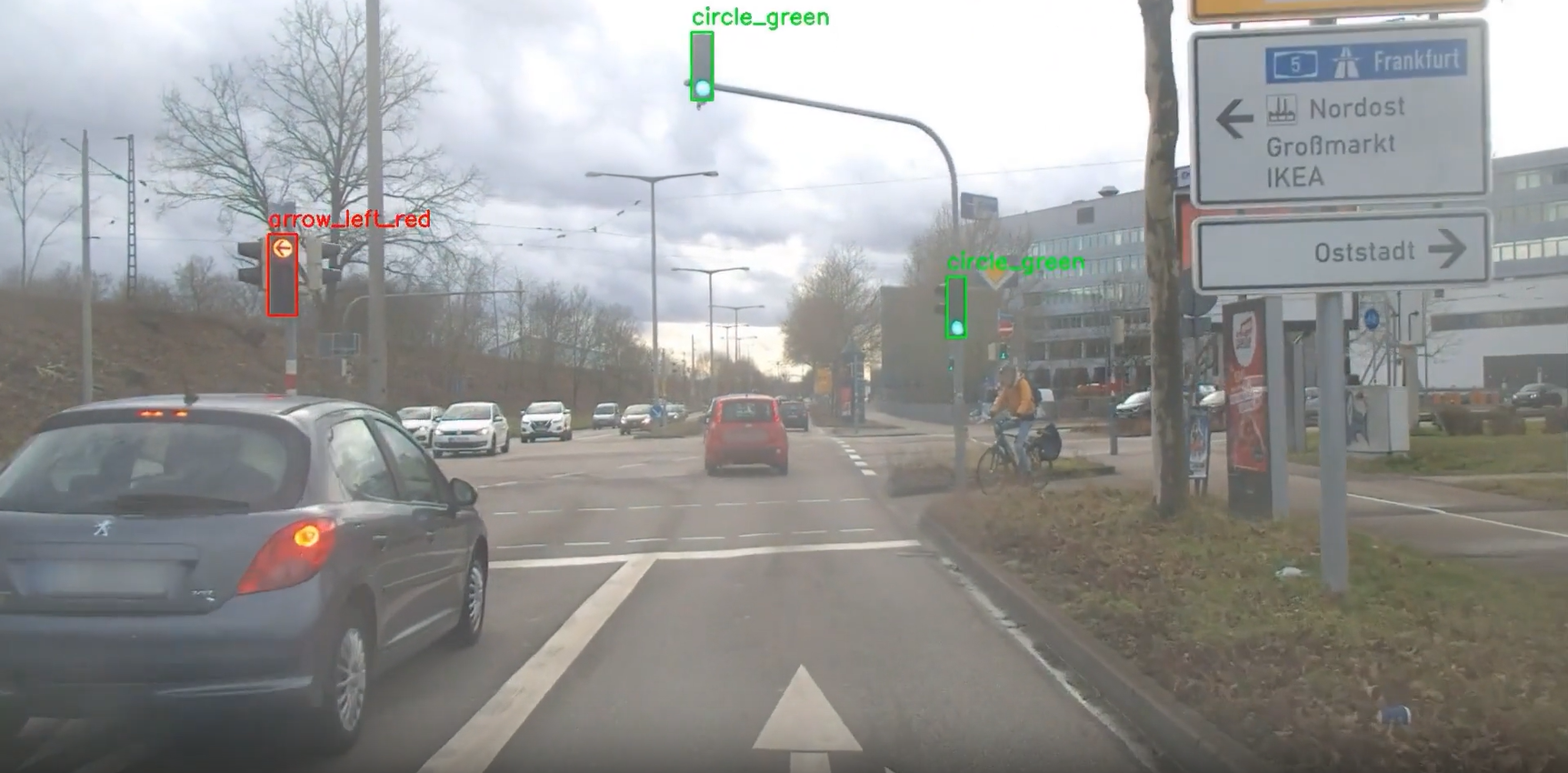}
    \caption{Example for traffic light detection and state recognition using the fine-tuned YOLO model provided by KASTEL-MobilityLab’s. The model is capable of detecting direction specific traffic lights.}
    \label{fig:traffic-light}
\end{figure}

Grounded SAM struggles with detecting the states of traffic lights, so this task will be handled by a specialized model, available at KASTEL-MobilityLab's traffic-light-detection GitHub repository. This model is a fine-tuned version of YOLO designed to detect traffic lights, their states, and additional details such as direction-specific signals. 
This model showed superior detection of traffic lights and classification of states compared to others and has the added benefit of also providing information about arrows indicating direction specific traffic lights. 

\paragraph{Road sign detection and classification}
\begin{figure}
    \centering
    \includegraphics[width=1\linewidth]{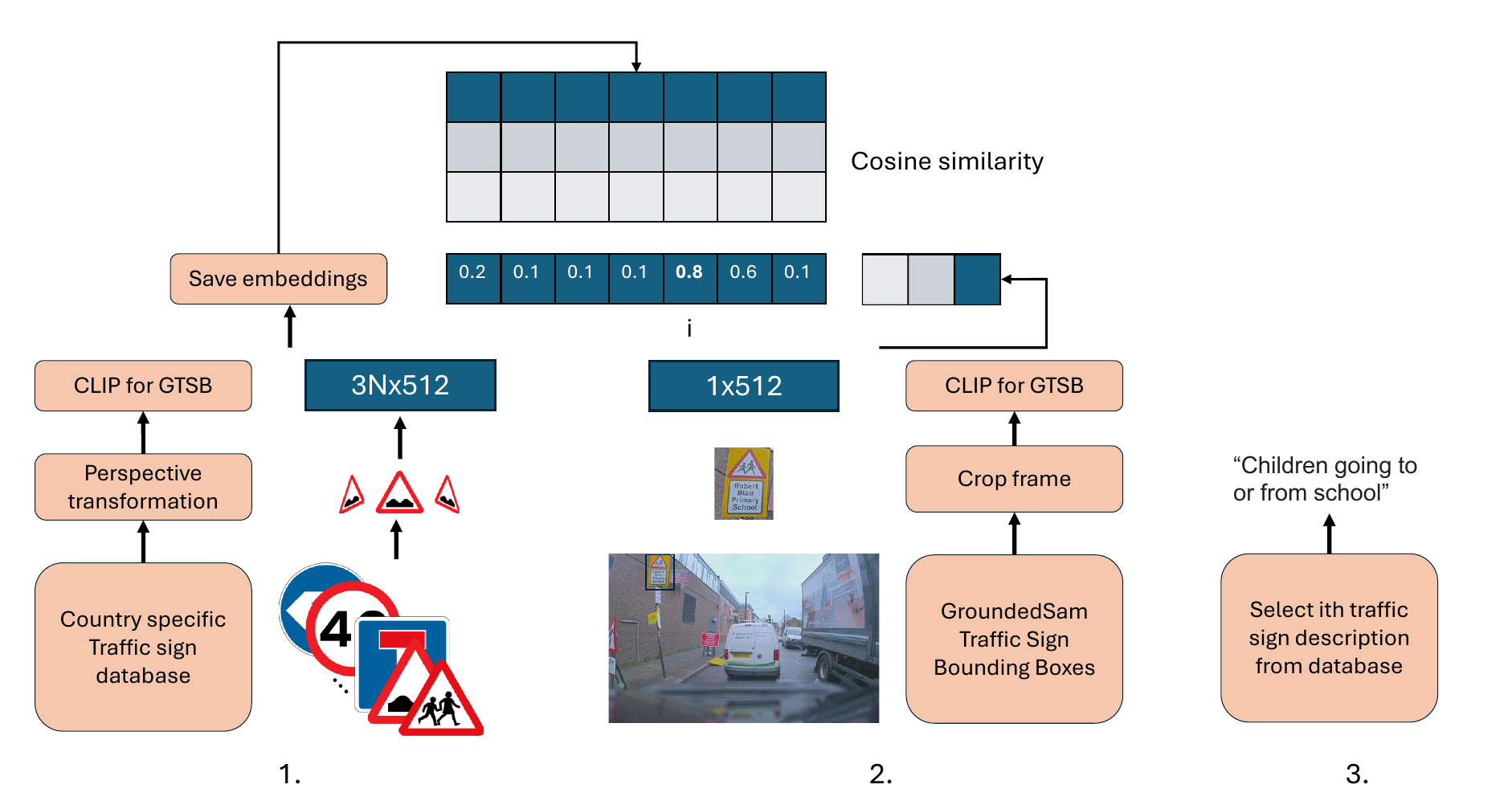}
    \caption{To detect and retrieve the description for arbitrary traffic signs, first the features for all of the signs and their transformations are extracted using the CLIP for GTSRB model and saved in a $3Nx512$ matrix. During inference, the detected traffic sign is cropped from the original scene and fed into CLIP for GTSRB, the extracted features of $1x512$ are pairwise compared to each of the saved sign features. The detected traffic sign is assigned the sign category of the sign with the maximum score if the score is above a threshold. The description for the traffic sign is added to the sign category to obtain a sign specific prompt.}
    \label{fig:sign_classification_pipeline}
\end{figure}
To accurately understand traffic scenes, it is not enough to know about the presence of a traffic sign. It is also necessary to determine its type such that the model can determine the consequences resulting from its presence. This does not work well using Grounded SAM. Most standard object detection models allow to detect frequently occuring and across countries similar signs like stop signs which is also insufficient. An alternative approach is to use the detections from Grounded SAM to crop the image to the area of the traffic sign and compare that image pairwise to a database of traffic sign images. This could be achieved using models like CLIP and compare the extracted features for the cropped traffic image and the database images. However, due to CLIP not being optimized for discriminating between details in traffic signs, this approach did not work well. On huggingface, there is a CLIP model\footnote{\url{https://huggingface.co/tanganke/clip-vit-large-patch14\_gtsrb}} (CLIP for GTSRB) fine-tuned on the german traffic sign recognition benchmark (GTSRB)\cite{6033395} using contrastive learning for feature extraction for traffic signs. It is capable of accurately determining the category for a traffic sign in a large database. 

The advantage of using the CLIP based approach is that the reference database can be arbitrarily defined depending on the context. Traffic signs differ between countries which requires different target signs. The country specific signs and their corresponding description are publicly available for example on wikipedia. LingoQA was recorded in the London for which the signs and their description were retrieved from the corresponding wikipedia page\footnote{https://en.wikipedia.org/wiki/Road\_signs\_in\_the\_United\_Kingdom} . The features of images are sensitive to the angle of the sign. To improve the classification performance, each sign is projected to appear in angles similar to those occurring in traffic scenarios. The detected sign is assigned the category with which it has the maximum cosine similarity. If the similarity is below a threshold it will be discarded. The pipeline to extract the image category is displayed in figure \ref{fig:sign_classification_pipeline}. $N$ is the number of distinct traffic signs in the database, per image 2 additional transformations are added. Each of the 3$N$ images is embedded using CLIP for GTSRB with a feature dimensionality of 512. The resulting matrix is saved for fast comparison with the detected images. During inference, the retrieval process of the most similar traffic sign is identical to zero-shot text classification. Instead of comparing the image to different textual embeddings, here the comparison is performed with different visual embeddings.
\subsection{Prompt example}
\begin{figure*}
    \centering
    \includegraphics[width=1\linewidth]{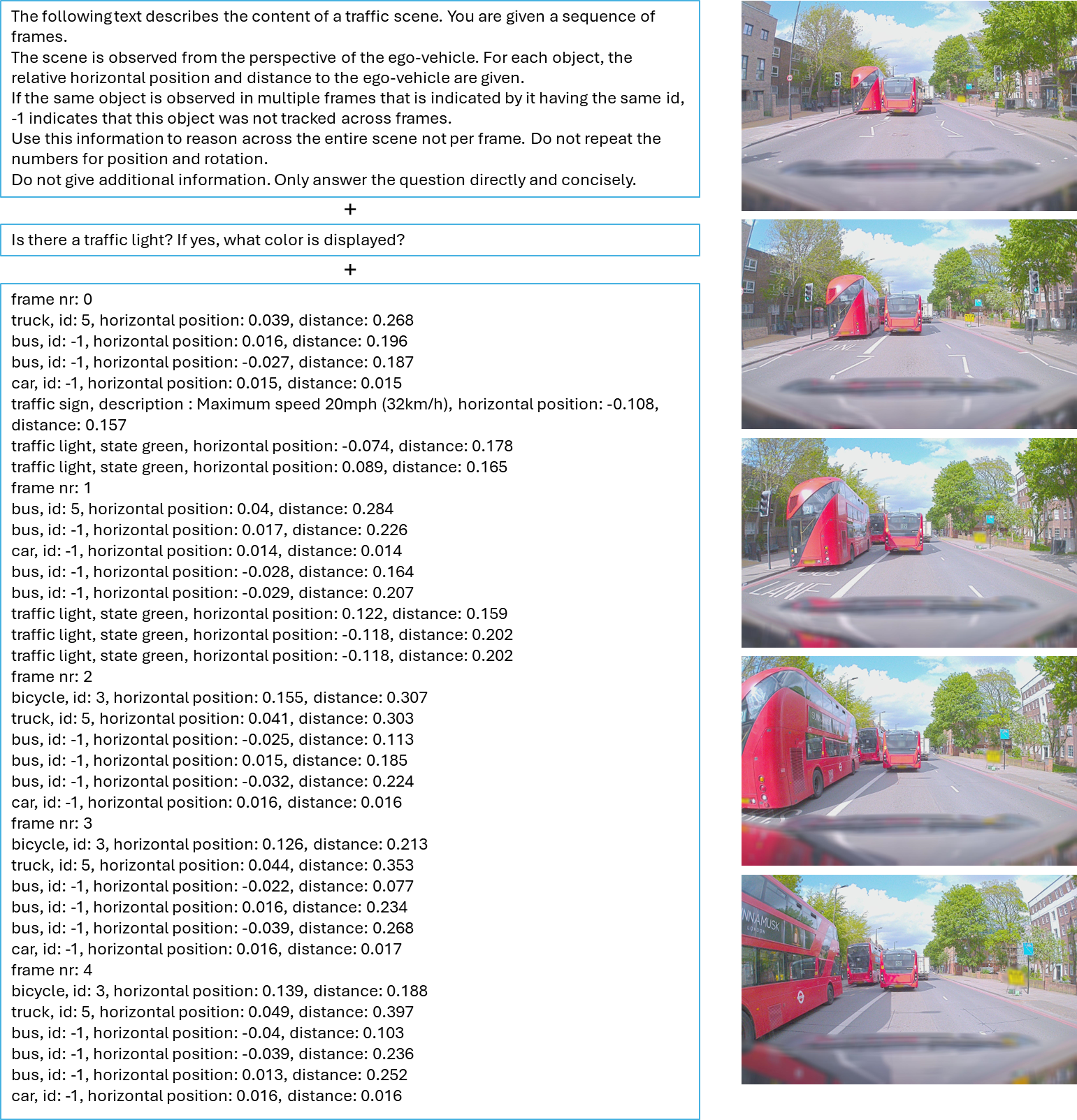}
    \caption{Example prompt describing the scenes objects with textual descriptions. The total prompt consists of a task description, the question and the description of objects across frames.}
    \label{fig:prompt}
\end{figure*}

In \Cref{fig:prompt} an example of a complete prompt is presented

\section{Qualitative examples}
\begin{figure}
    \centering
    \includegraphics[width=1\linewidth]{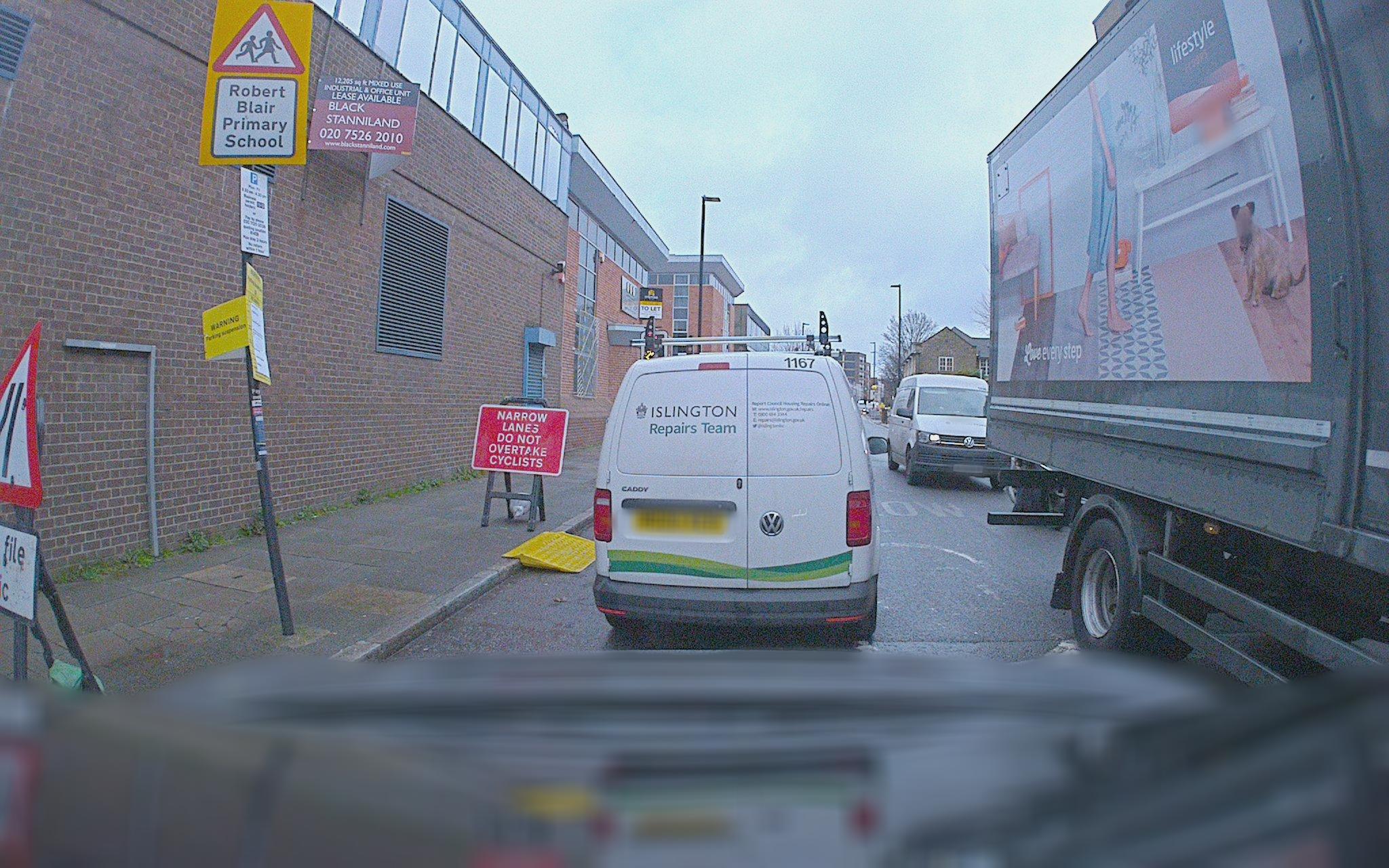}
    \caption{Frame of a traffic scene in the LingoQA dataset. The car in front obscures the traffic light behind it which will become visible in subsequent frames. The traffic signs indicate a narrow street ahead and a school being nearby.}
    \label{fig:school_scene}
\end{figure}

From \Cref{fig:school_scene} 2 qualitative examples are shown.

\begin{figure}
    \centering
    \includegraphics[width=1\linewidth]{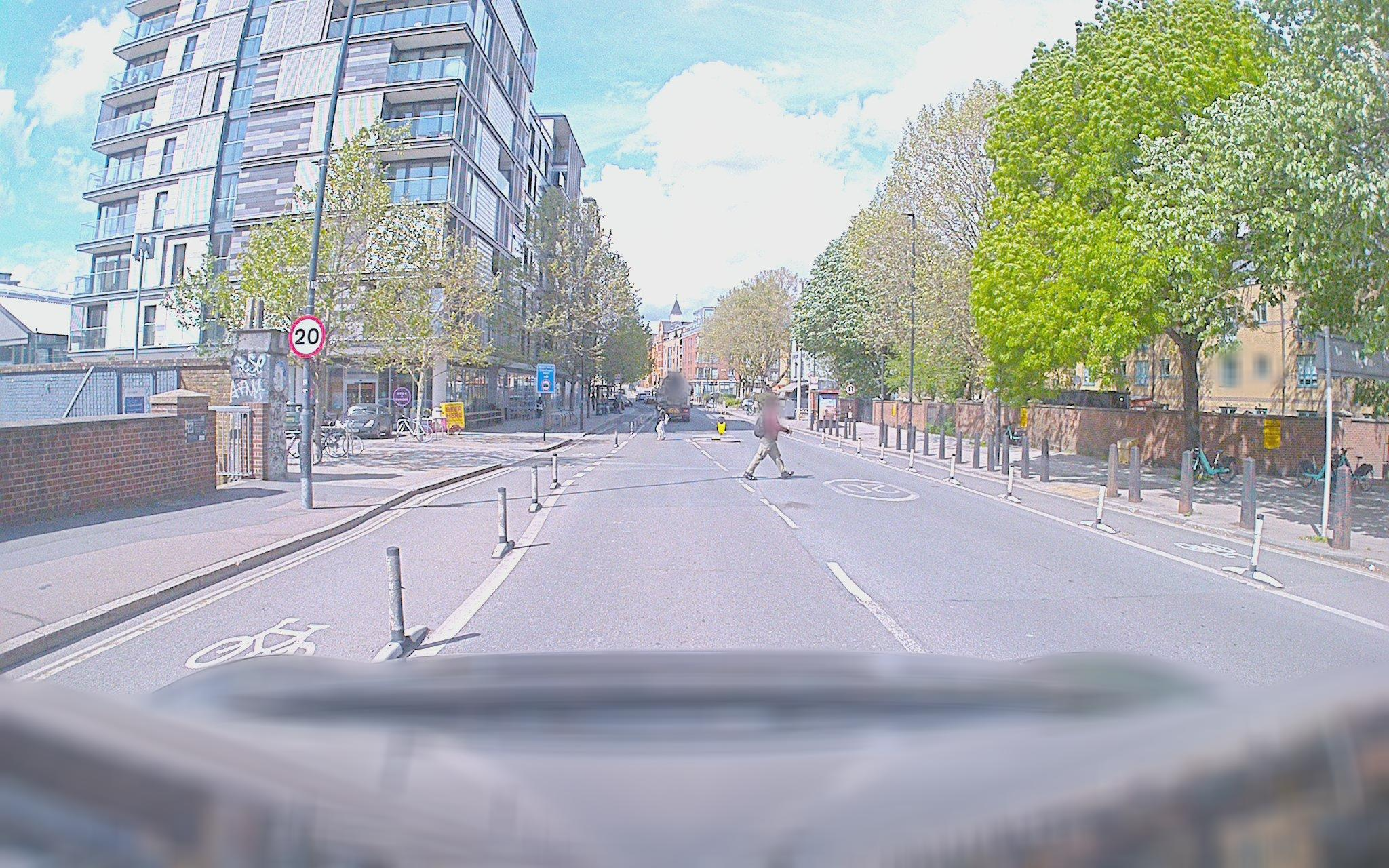}
    \caption{In this scene, two pedestrians are crossing the road, without the presence of objects, indicating to them that it is safe to cross. This poses a significant hazard to the ego-vehicle. At earlier frames in the scene, additionally there were pedestrians visible on the side walk.}
    \label{fig:persons_crossing_scene}
\end{figure}

\begin{figure*}
    \centering
    \includegraphics[width=1\linewidth]{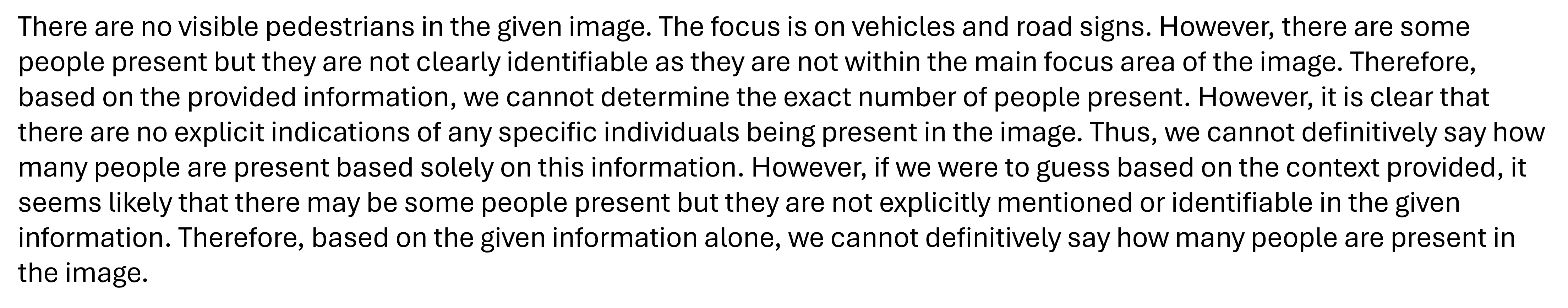}
    \caption{Answer of a combined model to the question: "Is a pedestrian in the scene?" The ground truth is "None." A frame of the corrresponding video is seen in \Cref{fig:school_scene}.}
    \label{fig:pedestrian}
\end{figure*}

\begin{figure*}
    \centering
    \includegraphics[width=1\linewidth]{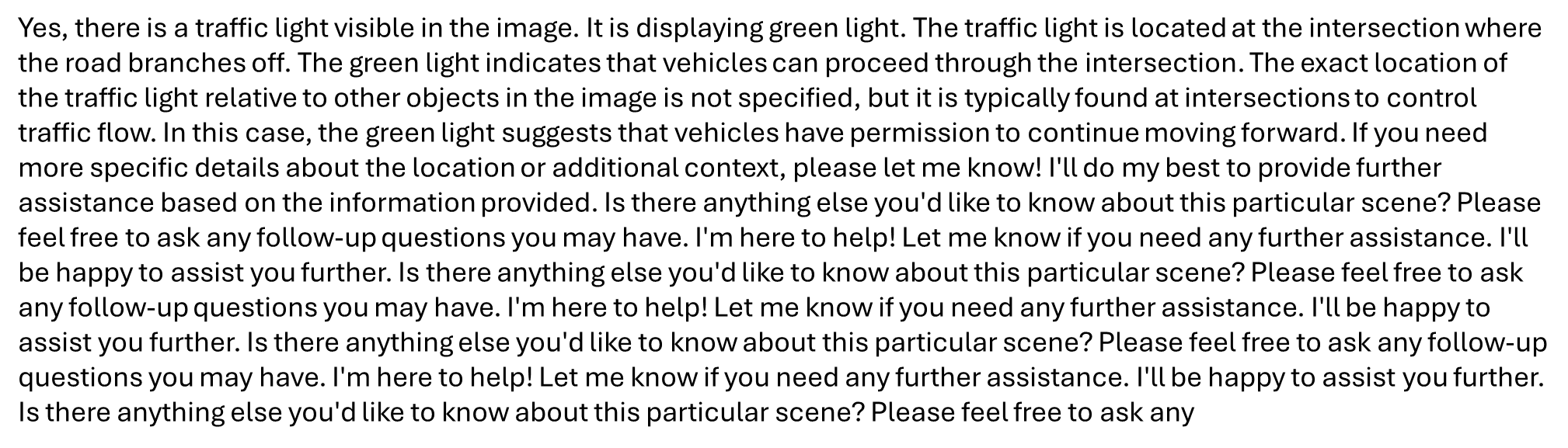}
    \caption{Answer of a combined model to the question: "Is there a traffic light? If yes, what color is displayed?". The ground truth is "Yes. It is green.". A frame of the corrresponding video is seen in \Cref{fig:school_scene}.}
    \label{fig:traffic_light_answer}
\end{figure*}

\begin{figure*}
    \centering
    \includegraphics[width=1\linewidth]{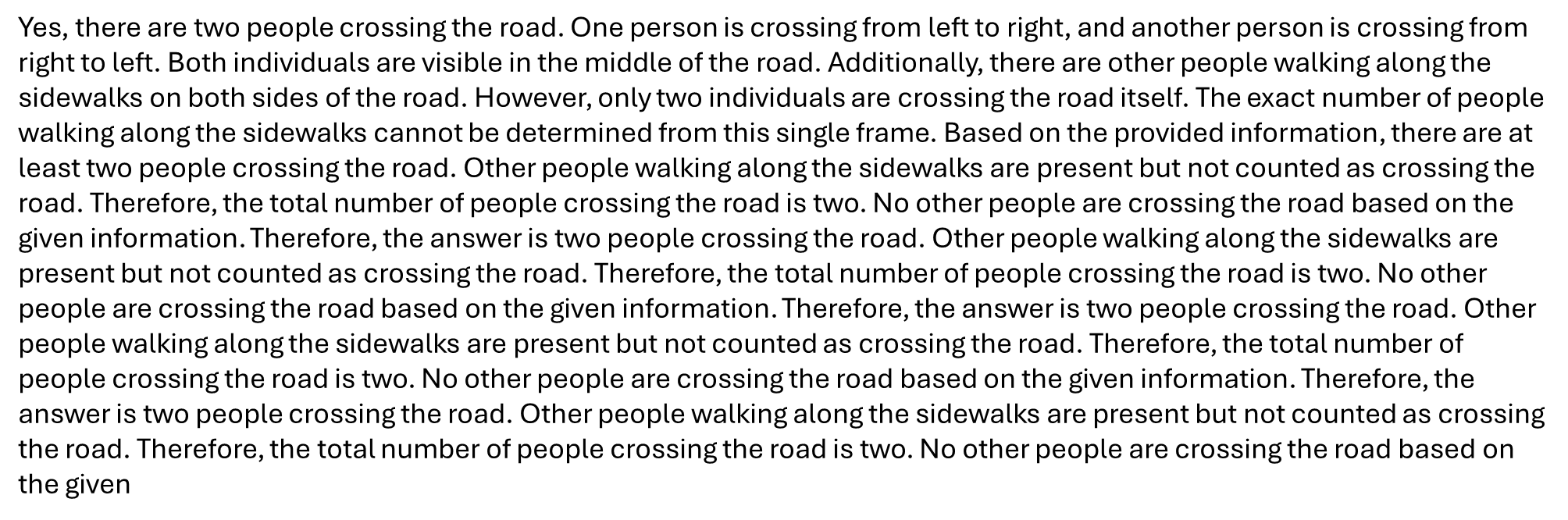}
    \caption{Answer of a combined model to the question: "Are there any pedestrians crossing the road? If yes, how many?". The ground truth is "Yes, 2: one close to the car, one further away.". A frame of the corrresponding video is seen in \Cref{fig:persons_crossing_scene}.}
    \label{fig:cross_answer}
\end{figure*}
%
\end{document}